\documentclass{article}
\pdfoutput=1 
\usepackage{arxiv}

\usepackage[utf8]{inputenc} 
\usepackage[T1]{fontenc}    
\usepackage{hyperref}       
\usepackage{url}            
\usepackage{booktabs}       
\usepackage{amsfonts}       
\usepackage{nicefrac}       
\usepackage{microtype}      
\usepackage{lipsum}
\usepackage{graphicx}
\graphicspath{ {./images/} }

\title{Evaluation of Seismic Artificial Intelligence with Uncertainty}

\author{
 Samuel Myren \\
  Los Alamos National Laboratory | Virginia Tech \\
  Los Alamos, NM | Blacksburg, VA \\
  \texttt{myrenst@lanl.gov} \\
   \And
 Nidhi Parikh \\
  Los Alamos National Laboratory \\
  Los Alamos, NM \\
   \And
 Rosalyn Rael \\
  Los Alamos National Laboratory \\
  Los Alamos, NM \\
   \And
 Garrison Flynn \\
  Los Alamos National Laboratory \\
  Los Alamos, NM \\
   \And
 Dave Higdon \\
  Virginia Tech \\
  Blacksburg, VA \\
   \And
 Emily Casleton \\
  Los Alamos National Laboratory \\
  Los Alamos, NM \\
}

\begin{document}

\maketitle
\section{Abstract}
    Artificial intelligence has transformed the seismic community with deep learning models (DLMs) that are trained to complete specific tasks within workflows. However, there is still lack of robust evaluation frameworks for evaluating and comparing DLMs. We address this gap by designing an evaluation framework that jointly incorporates two crucial aspects: performance uncertainty and learning efficiency. To target these aspects, we meticulously construct the training, validation, and test splits using a clustering method tailored to seismic data and enact an expansive training design to segregate performance uncertainty arising from stochastic training processes and random data sampling. The framework's ability to guard against misleading declarations of model superiority is demonstrated through evaluation of PhaseNet \cite{zhu_phasenet_2018}, a popular seismic phase picking DLM, under 3 training approaches. Our framework helps practitioners choose the best model for their problem and set performance expectations by explicitly analyzing model performance with uncertainty at varying budgets of training data. 
    
    \textbf{Keywords:} uncertainty quantification, model benchmark, deep learning, learning efficiency, phase pick, signal processing

\section{Introduction}    
    As seismology has breached the big data world within the last 20 years \cite{arrowsmith_big_2022}, the automation of seismic tasks through artificial intelligence (AI) via deep learning models (DLMs) is booming \cite{yu_deep_2021,mousavi_deep-learning_2022,mousavi_applications_2024}. However, different published models are often trained and evaluated using datasets that differ on important characteristics, making it hard to compare them directly. While growing availability of open-source benchmark data sets and software packages, such as SeisBench \cite{woollam_seisbenchtoolbox_2022}, has enabled researchers to compare popular models on the same datasets \cite{munchmeyer_which_2022}, we still lack robust evaluation methods to fairly evaluate and compare DLMs. Without a robust evaluation approach, we are unable to determine whether new technological developments are truly state of the art, we lack diagnostics that assist developers in pinpointing areas for model improvement, and we fail to convey the scenarios under which a model can be trusted. In this research, we address this gap by identifying two fundamental evaluation aspects for DLMs and uniting them under a single evaluation framework.
    
    The first and perhaps most important evaluation aspect is \textbf{performance uncertainty} in terms of benchmarking. Two identical DLM architectures with identical training approaches will realize different performance metrics when tested on a benchmark dataset as a result of different initializations and random batching during training \cite{bouthillier_accounting_2021,reimers_reporting_2017}. We refer to this as training uncertainty. Furthermore, the training data sample has drastic random effects on performance metrics \cite{bouthillier_accounting_2021}. We refer to this as data uncertainty. Determining algorithmic superiority in light of both sources of uncertainty is a known issue \cite{goldstein_league_1996, dietterich_approximate_1998, dehghani_benchmark_2021, wein_follow_2023} and, to our knowledge, has not been explored for the seismic AI community. We note that our aim is not to quantify \textit{prediction uncertainty}, which entails relating model variability in predictions to the prediction errors quantitatively, typically using statistical confidence/credible intervals.
    
    The second vital, yet often unexplored, evaluation aspect is the model's \textbf{learning efficiency} in terms of data utilization. Adding redundant data (too many similar examples) does not necessarily add new information, making model training inefficient. So, here, learning efficiency is defined as the relationship between the gain in performance as a result of the additional training data and data diversity required (see also \cite{hlynsson_measuring_2019}). Understanding a model's learning efficiency helps decision makers compute the cost to benefit ratio of acquiring additional and more diverse data to meet a problem's performance requirements, especially in data scarce scenarios \cite{alzubaidi_survey_2023}. It is particularly important in the context of transfer learning (TL) where models trained in one context are fine-tuned with smaller, targeted datasets for another context \cite{bornstein_pickblue_2024}, and the models that best adapt are preferred. 

    Measuring learning efficiency while accounting for uncertainty may appear to be simple at first. One might train two models on different budgets of randomly sampled training data, compare their performances, and repeat under re-sampled data to account for data uncertainty. However, randomly sampling data does not take correlations between seismic data into account. Waveforms may share similar signal properties if they originate from similar sources, come from nearby stations, or follow similar paths. This is a form of data leakage \cite{ferrerflorensa_spanseq_2024}, where one dataset contains overlapping information from another dataset. Data leakage can influence all aspects of model evaluation. For example, when identical or nearly identical data points exist in both training and test datasets, model performance can be overestimated. Failing to account for data leakage when choosing training budgets can lead to inaccurate measurements of learning efficiency. This is because as more data is added, there is no control over how much new information (or \textit{diversity}) is being added. That is, without controlling for data leakage in developing training budgets, one may fail to systematically expose the models to more data diversity. Finally, failure to account for it can induce unrealistic representation of uncertainty arising from data sampling, potentially underestimating its impact by downplaying bias in the sampling procedure. In this work, we mitigate data leakage to more directly measure learning efficiency and better represent realistic data uncertainty by using stratified sampling to construct the training data budgets.
    
    This research develops an evaluation framework for seismic DLMs that considers two aspects: performance uncertainty arising from data and training, and learning efficiency. We utilize concepts from statistical experimental design to measure each aspect in turn. We demonstrate our framework using a seismic DLM to directly compare 3 training approaches: training from scratch and two TL approaches. We measure uncertainty through deep ensembles and random data selection processes and learning efficiency through careful data splitting and varying data amounts. Finally, we cast all performance results into a statistical framework for direct comparison with uncertainty.
    
    We use a popular seismic DLM for demonstrating the evaluation framework. A plethora of research in applying DLMs to the cardinal task of phase-picking provides a strong basis for our demonstration. Phase picking is the task of identifying the time that event related signatures arrive, typically the primary and/or secondary phases (P-arrival and S-arrival, respectively), at a seismic station. Phase picking is a vital task for detecting, locating, and characterizing seismic events such as earthquakes or explosions. It can be a cumbersome task, motivating researchers to actively construct better DL phase picking models (referred to as pickers hereafter) \cite{ross_generalized_2018, mousavi_earthquake_2020, chen_cubenet_2022, zhu_phasenet_2018, zhu_deep_2019, zhu_endend_2022, tokuda_seismic-phase_2023, bornstein_pickblue_2024, niksejel_obstransformer_2024} and others are using these pickers to expedite workflows and create catalogs \cite{cianetti_comparison_2021, park_basement_2022,yoon_detailed_2023, armstrong_deep-learning_2023}. Of course the best performing and most robust picker is preferred, but determining which picker is best for a given context remains a challenge that our framework is positioned to address. 

    The paper is formulated as follows. We first introduce the data sets and how the data splits are constructed to incorporate the two fundamental evaluation aspects. Then we briefly introduce the model architecture and training scheme, including curating the inputs and targets, and provide a detailed review of the 3 training approaches used for comparison. This is followed by the evaluation approach including the statistical framework that distills the different model performances for digestible comparison with uncertainty. Finally, we present the results in terms of performance and uncertainty and finish with concluding remarks and future work.

\section{Data}

    \begin{figure}
        \centering
        \includegraphics[]{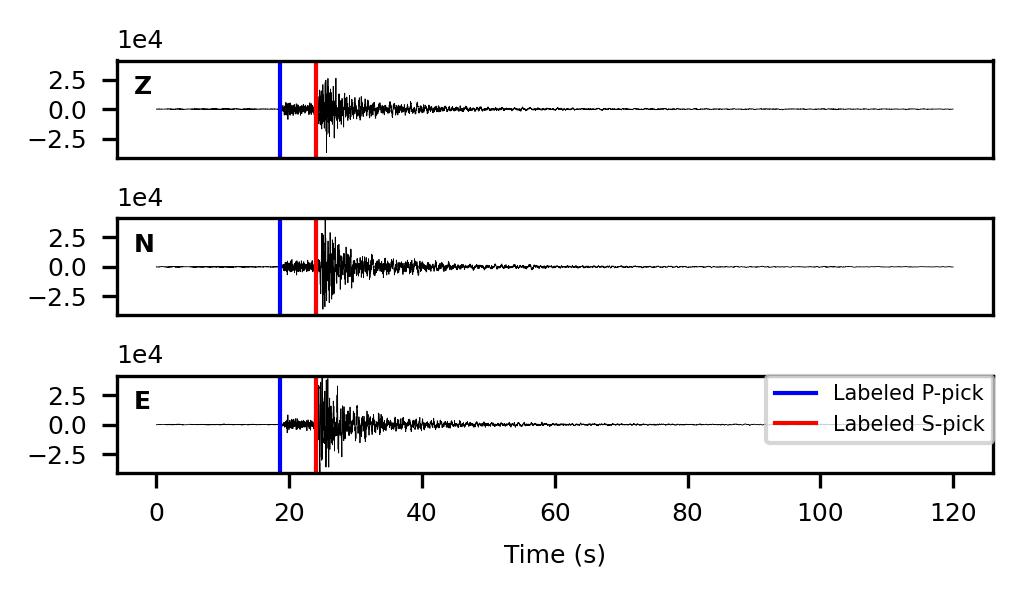}
        \caption{An example waveform from the INSTANCE dataset showing the 3-component response in counts (Z - top, N - middle, E - bottom) with the labeled P-arrival (vertical blue line) and labeled S-arrival (vertical red line).}
        \label{wf}
    \end{figure}

    We use two publicly available, labeled, non-continuous waveform datasets: the Stanford Earthquake Dataset (STEAD; \cite{mousavi_stanford_2019}) and the Italian seismic dataset for machine learning (INSTANCE; \cite{michelini_instance_2021}), both accessed through the Python seismic deep learning package, \texttt{SeisBench}. STEAD is used for pre-training and INSTANCE for subsequent fine-tuning and testing. STEAD contains $\sim235k$ noise waveforms and $\sim1.03M$ earthquake waveforms representing $\sim441k$ unique sources. For the INSTANCE dataset, we use the version measured in counts (\texttt{InstanceCountsCombined} in \texttt{SeisBench} documentation). In addition, we use the instrument response deconvolved traces in velocity without redundant measurements in acceleration. The resulting INSTANCE set includes $\sim132k$ noise waveforms and $\sim824k$ earthquake waveforms arising from $\sim54k$ unique sources. An example waveform from INSTANCE is shown in Figure \ref{wf} along with the labeled P-arrival and S-arrival.

    \subsection{Training, Validation, and Test Splits}

        We use the default splits provided in \texttt{SeisBench} for STEAD for pre-training the models, which we then fine-tune using INSTANCE data. We implement a more meticulous splitting process that is meant to reduce data leakage for INSTANCE, since we use it to measure learning efficiency with uncertainty. We construct the splits using stratified sampling on source locations. We chose location as a stratifying feature because it was an easy-to-implement first order way to segregate the data such that it captures some diversity and represents how seismic datasets are often constrained to specific geographic regions, thereby roughly mimicking a realistic data sampling procedure. For example, many of the publicly available benchmark datasets in \texttt{SeisBench} are sourced from specific geographical regions. In practice, one might want to consider how model performance improves with the inclusion of data from new regions, or how a model might perform when predicting activity in a new region with limited data. 

\begin{figure}
            \centering
            \includegraphics[]{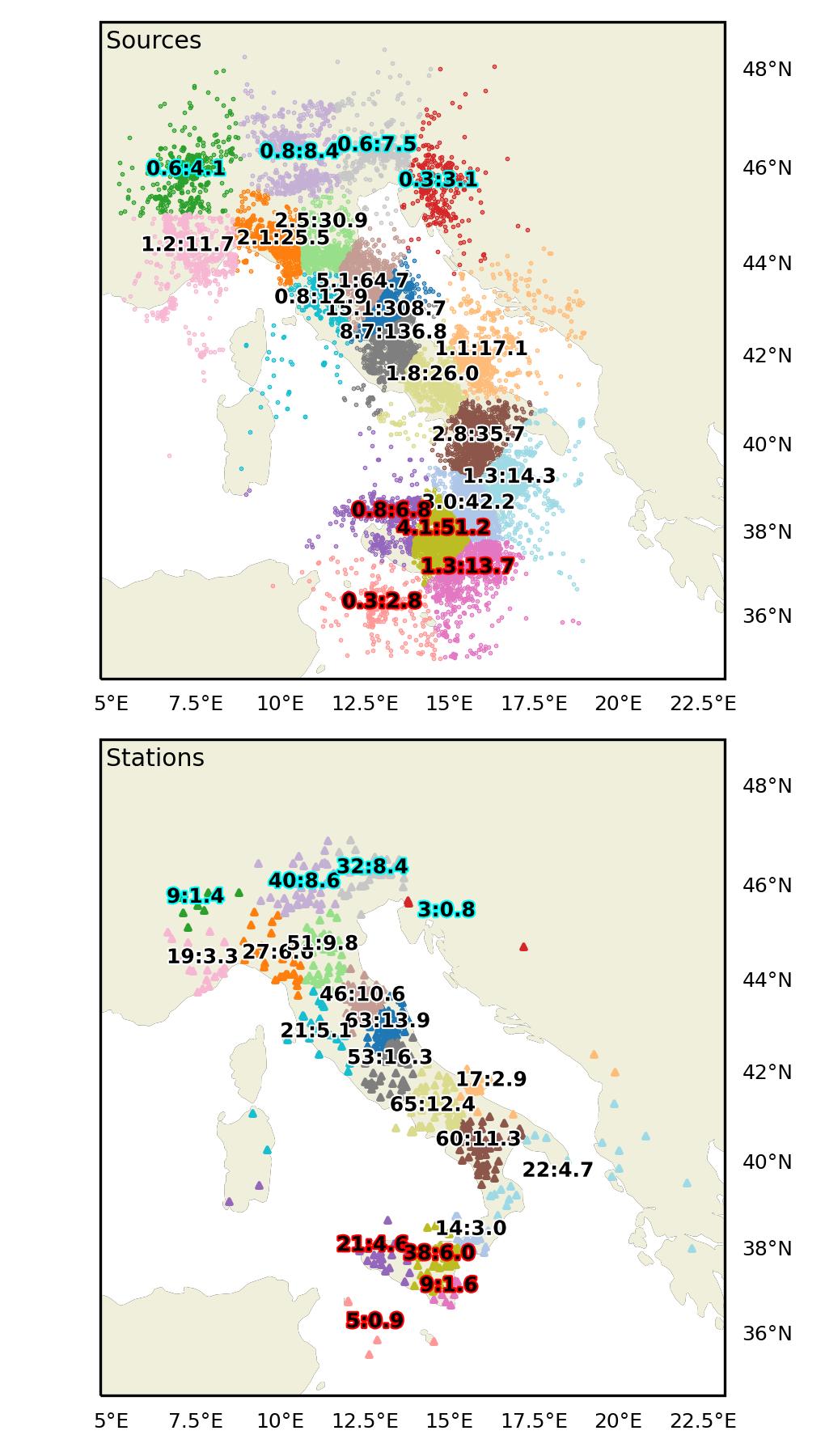}
            \caption{INSTANCE data with 20 colored clusters for the sources (top, circles) and the stations (bottom, triangles). For the sources, the numbers indicate the number of sources in thousands followed by the number of waveforms in thousands. For the stations, the numbers show the number of stations and the number of noise waveforms in thousands. The colored text outline designates the group: north test group is cyan, training/validation group is white, and south test group is red.}
            \label{clusters}
        \end{figure}

        To accomplish our stratification, we cluster the INSTANCE sources spatially based on their latitude and longitude using k-means algorithm from the Python package, \texttt{Scikit-learn} \cite{pedregosa_scikit-learn_2018}. K-means is an iterative, centroid based clustering algorithm. Initially all cluster centroids are selected at random and all data points are assigned to a cluster closest to them. Next, in each iteration, a new centroid is estimated for each cluster based on the mean of all points assigned to this cluster and then these new centroids are used to assign clusters for each point (i.e., based on the nearest cluster centroid). This process continues until convergence. We choose 20 clusters because it provided clusters large enough to feasibly represent the region's activity, yet small enough to explore model performance in limited data settings. To assign the noise waveforms, which have no corresponding source locations, to the established clusters, we use the trained k-means model to predict its associated cluster based on its station location. The clusters and number of sources, stations and waveforms are shown in Figure \ref{clusters}. Table \ref{table:splits} summarizes the amount of data in the training, validation, and test splits and the construction of the splits are described next. 
        
        To construct the INSTANCE splits, we first segregate the clusters into two groups: a training/validation pool and a test pool. We select the 12 central clusters to be the pool from which the training/validation data will be constructed, and 8 clusters - the 4 northern most and 4 southern most - to be the test clusters (see Figure \ref{clusters}). Choosing the test clusters from northern and southern regions allows us to explore two different concepts. First, they help to evaluate model performance on semi-out-of-distribution data. Second, these splits represent a situation where we may know how a model performs in a certain region and we want to measure its consistency for neighboring regions. 

        To mitigate regional bias in the the INSTANCE test split, we balance the data from north and south regions. The northern region is smaller comprising 2,269 sources, so we randomly sample 2,269 sources from the southern region. For each source selected in the test split, we include all associated waveforms, resulting in a test set totaling 49,516 earthquake waveforms: 23,086 from the north and 26,430 from the south. To maintain the overall ratio of 11.4\% noise to earthquake waveform ratio originally in INSTANCE, we randomly select 2,822 noise waveforms from each of the south and north regions. 

        \begin{table}
            \caption{The data splits are summarized showing the number of sources, earthquake (EQ) waveforms (WFs), and noise WFs. For the training split, values should by multiplied by the quantity of clusters used in the training design.}
            \centering
            \begin{tabular}{llll}
            \toprule
            \textbf{Data Split}   & \textbf{Sources} & \textbf{EQ WFs} & \textbf{Noise WFs} \\
            Training              & 636 $\times$ \# of clusters   & 8,878 $\times$ \# of clusters & 1,106 $\times$ \# of clusters  \\
            Validation            & 1,908                         & 27,169                        & 3,312              \\
            Test                  & 4,538                         & 49,516                        & 5,644              \\
            \bottomrule
            \end{tabular}
            \label{table:splits}
        \end{table}

        We aim for an 80-20 ratio for the training to validation splits and we desire an equal number of sources from each cluster to reduce bias. Since the smallest of the 12 central clusters has 795 sources, we randomly sample 159 sources from each cluster for the validation set. This totals to 1,908 sources encompassing 27,169 earthquake waveforms. Maintaining a $\sim11.4\%$ earthquake to noise ratio, we then randomly sample 276 noise waveforms per cluster totaling 3,312 noise waveforms for the validation split. 
    
        After removing the INSTANCE validation set from the central clusters, the remaining waveforms become the pool from which we build the training splits. When training models, the amount of training data is specified by the number of clusters used for training. However, to keep data amounts relatively constant across clusters, each time a cluster is selected to be included in training data, we randomly select 636 sources (80\% of 795, the smallest number of sources in any cluster) which on average results in 8,878 earthquake waveforms, and 1,106 noise waveforms per cluster.
        
\begin{figure} %
            \centering
            \includegraphics[]{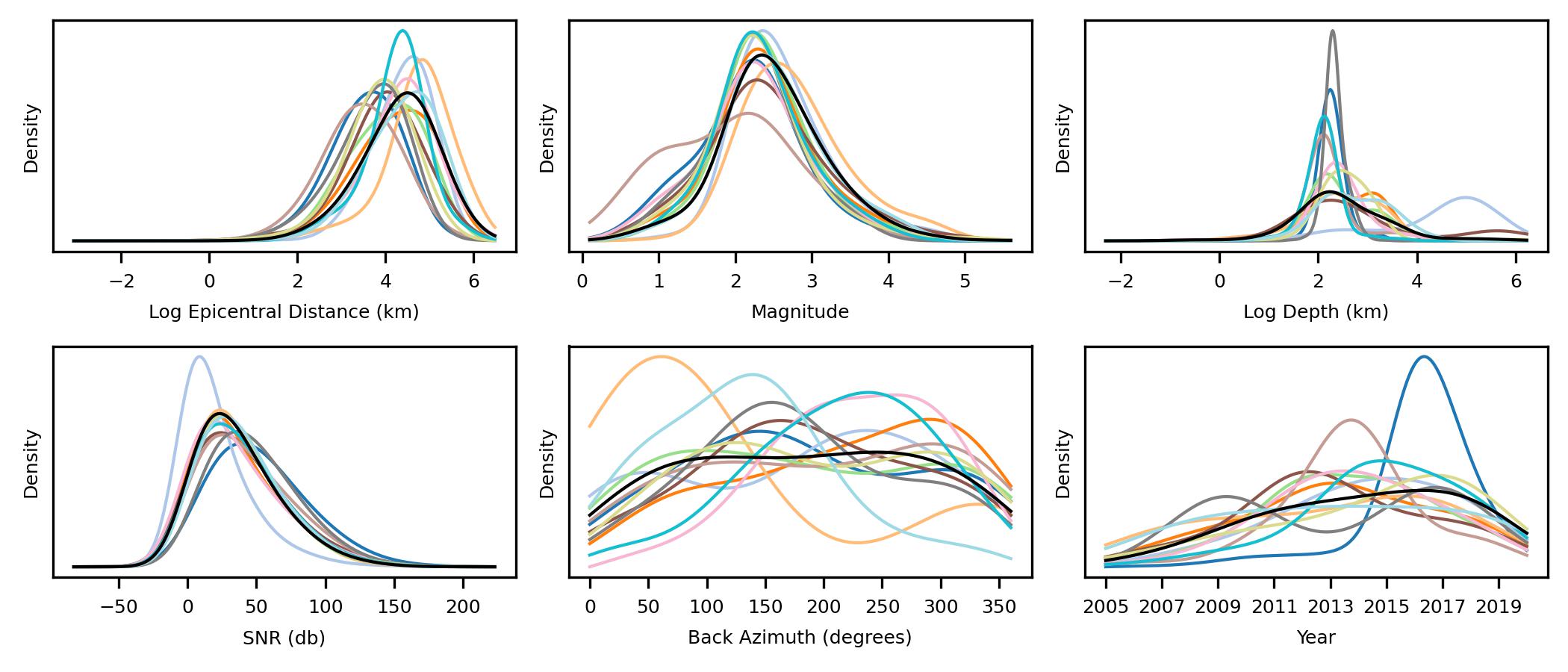}
            \caption{Distributions of selected data features (log epicentral distance, magnitude, log depth, SNR averaged over the 3 components, back azimuth, and source year) for each of the 12 training clusters (non-black curves corresponding to Figure \ref{clusters}) and the testing set comprising the 4 northern and 4 southern clusters (black curve). These features are provided as metadata in the INSTANCE dataset found in \texttt{SeisBench}. All curves are normalized for comparison such that they integrate to 1.}
            \label{cluster_features}
        \end{figure}

        In our stratification, our aim was to choose a realistic feature that captures data diversity, mimics data sampling variability, and mitigates data leakage and thus reduces bias in measuring learning efficiency. We acknowledge that there are many other features that seismologists may use to stratify the waveform data, including time, network, geological faults or other geological features, source-station distance, magnitude, instrument type, earthquake type, signal to noise ratio (SNR), signal characteristics, and possibly many others. We also recognize that partitioning data on any single feature does not fully capture all forms of data diversity. However, we find that partitioning on source locations induces some segregation in these other data features. Figure \ref{cluster_features} provides deeper insight into the how the geographical clustering impacts the distributions of selected metadata features derived from the INSTANCE dataset. More details and figures showing distributions of other data features and waveform signal properties of the clusters are provided in the appendix (Figures A1-A5). Other details regarding INSTANCE data, including geographical visualizations of source focal mechanisms and magnitudes, can be found in \cite{michelini_instance_2021}. 
        
\section{Methods - Model Architecture \& 3 Training Approaches}

    \begin{table}
        \caption{The details of the 3 models are given in each row including their names, if the model is initialized randomly or using weights from an upstream model trained on STEAD, and whether weights were frozen.}
        \centering
        \begin{tabular}{lll} %
        \toprule
        \textbf{Name}  & \textbf{Initialization} & \textbf{Weights}  \\
        Standard          & Random        & Free         \\
        TL Free           & STEAD         & Free         \\
        TL Frozen         & STEAD        & Half Frozen  \\
        \bottomrule
        \end{tabular}
        \label{table:mods}
    \end{table}

    We develop a unifying framework for evaluating AI models that considers performance uncertainty and learning efficiency to guide model selection. We demonstrate our framework by comparing 3 training approaches outlined in Table \ref{table:mods}: training from scratch (the standard DL approach) and two TL approaches with varying number of free parameters. However, the framework presented is sufficiently broad where one could explore the impacts of any aspect of interest, such as different hyperparameters, training approaches, or new architectures altogether. We first review the model architecture and other commonalities among the models, then we review the uniqueness of each model and how the training design allows for estimation of data and training uncertainty. 

    All models use the PhaseNet deep learning architecture which has been extensively utilized for phase picking \cite{munchmeyer_which_2022,kim_development_2023, bornstein_pickblue_2024} and features a U-Net type architecture \cite{ronneberger-unet_2015}. PhaseNet intakes 3-component waveforms of dimension 3,001 and is trained to output probabilities of P-arrivals, S-arrivals, and noise for each time point in the waveform. 

    We curate the training inputs by applying a 30s window randomly around the 100 Hz sampled 3-component waveform. For earthquake waveforms, we require the window to include the labeled P-arrival. The windows are re-randomized every epoch. Then, each waveform component is independently zero-meaned and scaled to unit variance. Due to the data curation of STEAD and INSTANCE, the P-arrivals tend to appear early in the waveforms, most often within the first 15s. As such, the random windowing reduces some temporal bias, but some remains, and this is a source of bias induced by the training process. There has been some work inspecting how windowing and overlap can impact predictions \cite{pita-sllim_parametric_2023, park_basement_2022, park_making_2024}, but this is not a focus of this work. Furthermore, we recognize the challenge of waveforms containing multiple unlabeled arrivals and how data augmentation can help mitigate this issue, but no other data augmentation was implemented as our focus is in the evaluation framework as opposed to progressing picker technology.

    For the training targets, we follow a process similar to \cite{zhu_phasenet_2018}, where we construct the P and S training targets using a normal distribution centered at the labeled picks with a standard deviation of 0.1 seconds. For the noise component, we simply subtract the P and S target components from a vector of ones. Since our goal is to demonstrate the evaluation framework, we focus only on the P-arrival predictions and discard the S-arrival and noise predictions for all subsequent analysis. 
    
    To demonstrate our evaluation framework, we compare 3 training approaches: the standard approach for deep learning and two TL approaches. The standard approach entails randomly initializing model weights and pulling the training, validation, and test splits all from the dataset of interest, in this case INSTANCE. Thus, the Standard model represents this approach. 
    
    The two TL approaches are thought to be useful when a small amount of labeled data is available from the region of interest, but a large amount of labeled training data can be obtained from other regions. As such, one first trains a model with data from the other regions (we call an upstream model), then switches to data from the region of interest to continue learning, thereby transferring the accumulated knowledge to the new domain. In our case, we first train PhaseNet using STEAD and then fine-tune using INSTANCE. We explore two options. The first allows all weights to change during INSTANCE training phase as similarly performed in \cite{chai_using_2020}. The second fixes the weights of the input and convolution layers (i.e., the left half, or encoder-side, of the network). Previous work exploring optimal layer freezing for U-Net architectures for medical image segmentation demonstrates that this choice may be non-optimal \cite{amiri_fine-tuning_2020}, but our objective is model evaluation, not finding the optimal approach for TL. The TL models that have fully free weights are denoted as Free, and those with some fixed weights are denoted as Frozen. All 3 models (the Standard model and 2 TL models) are validated and tested using INSTANCE data. 

    To evaluate learning efficiency, we train each model using 1, 3, 6, 9, and 12 clusters from the INSTANCE training clusters. To measure data uncertainty at a given quantity of clusters, we randomly select the cluster set 12 times, training the model each time. Then, to separate the data uncertainty from training uncertainty, we use 4 deep ensembles \cite{lakshminarayanan_simple_2017,huang_snapshot_2017,ganaie_ensemble_2022}. That is, for each of the 12 cluster sets at a given quantity of clusters, we retrain the model 4 different times under different initializations, holding the training data sample fixed, to obtain 4 model instances for evaluation. To obtain different initializations for the Standard model, we simply randomly reset the weights. For the TL models which are seeded by weights from an upstream model, we first train 48 upstream models with STEAD starting from random initializations. Then, we initialize the TL models by pulling the set of weights from one randomly selected upstream model. 
    
    In our deep ensemble implementation, we loosely follow the snapshot ensemble approach of \cite{huang_snapshot_2017} which boasts computational advantages in finding different parameter states that minimize loss. We implement a cosine annealing learning schedule with 4 restarts with an initial learning rate of $0.01$ and final rate of $0$. These learning parameters exhibited the best performance in preliminary training experiments especially for models with the smallest quantities of clusters (one cluster). However, simple rate resets routinely failed to push models with small training sets into new basins of attraction, so we altered the approach towards the classical deep ensemble method \cite{lakshminarayanan_simple_2017} by fully reprogramming the model weights at each scheduled rate reset rather than relying on rate resets alone. Depending on the model being trained, this reprogramming may be a complete randomized reset (for the Standard model) or replacing weights by randomly pulling from one of the 48 upstream models (for the TL models). 
    
    For ease of discussion, we define a single "model instance" as the combination of model, quantity of clusters, cluster set (random selection of those clusters), and initialization (ensemble). For example, a single model instance of the Standard model could be trained using 3 clusters randomly chosen - in reference to Figure \ref{clusters} - to be clusters red, blue and light-green, and the third initialization. With the Standard model and 2 TL models each trained using 5 quantities of clusters over 12 random cluster sets with 4 initializations, we obtain 720 model instances. 

    We train all models with the deep learning Python package, \texttt{PyTorch} \cite{paszke_pytorch_2019} using a sufficient number of epochs to allow full convergence. The Standard models reached convergence within 50 epochs and the TL models reached convergence within 12 epochs. We selected the model instance for evaluation that exhibited lowest loss on the validation data within a cosine annealing cycle. We used cross entropy loss, the ADAM optimizer \cite{kingma_adam_2017}, and a constant batch size of 128 due to its superior performance in preliminary testing.

\section{Evaluation}
    
    \subsection{Metrics}
    
        As our main goal is to demonstrate the effectiveness and versatility of our evaluation framework, we limit the scope of evaluation of the pickers by displaying just a couple classification metrics and one functional based regression metric. In this section, we briefly outline how we calculate these metrics. To motivate the specific metrics we consider, we note that it is important in evaluation of pickers to consider both the model's strength in identifying if a P-arrival occurs (e.g., alerting) and, if so, the associated accuracy of the pick (e.g., how close is the predicted pick to the labeled P-arrival). 
        
        In general, we implement techniques similar to those found in \cite{zhu_phasenet_2018, munchmeyer_which_2022,kim_development_2023} to obtain model outputs and picks for each test waveform. More specifically, we feed test waveforms through the model in 30 second intervals with 28 seconds of overlap and no end-padding. Then, we take the mean of the model outputs at each time-point to form a single vector of P-arrival probabilities as a function of time for each waveform. Given a defined threshold (described later), we define a pick as the discrete locations in time that a P-arrival is predicted by the model by taking the argmax of groups of samples that contiguously exceed the threshold. 
        
\begin{figure}
            \centering
            \includegraphics[]{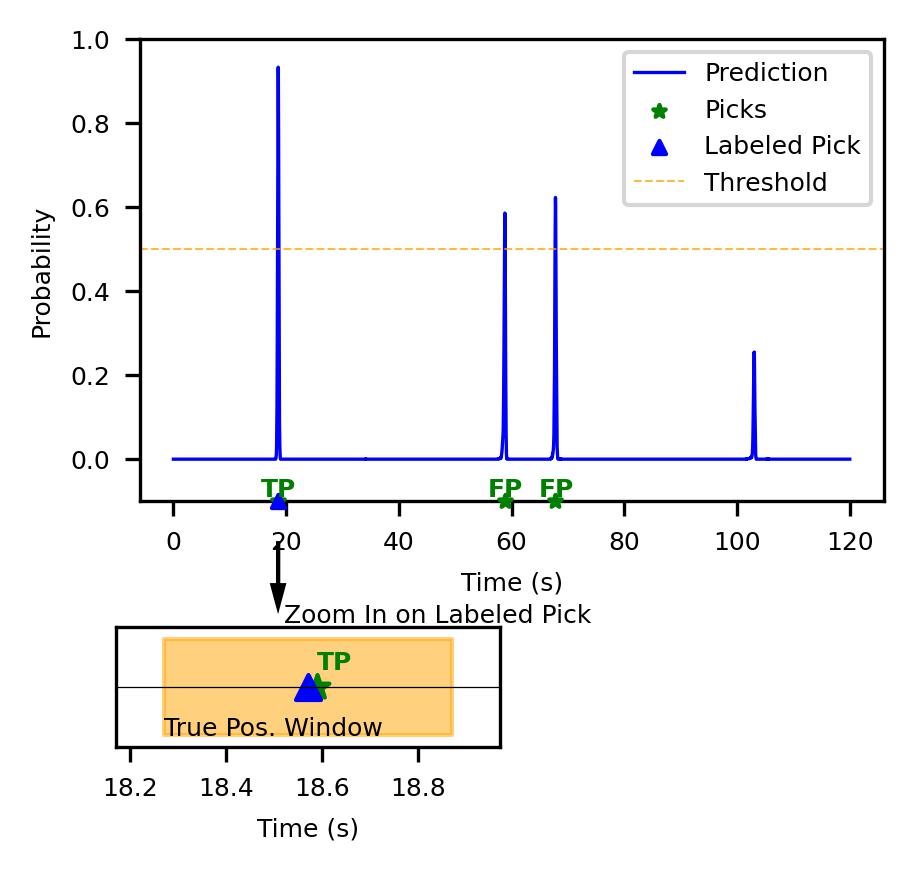}
            \caption{An example of a model output for a test waveform showing the predicted probability of the P-arrival at each time point. The labeled P-arrival is displayed on the x-axis as a blue triangle along with 3 predicted picks (green stars) and labels classifying the pick as either a FP or TP. An inset plot zooms in on the true positive window (orange box: $\pm 0.3$ seconds around the labeled P-arrival) to show the TP. Here, an arbitrary threshold of 0.5 (orange line) was used for example purposes.}
            \label{pred_example}
        \end{figure}

        Evaluating the model's alerting capability is typically done using classification metrics defined from the picks \cite{zhu_phasenet_2018, munchmeyer_which_2022,kim_development_2023, yu_benchmark_2023}. We define a pick as a true positive ($TP$) if it falls within a \textit{true positive window} of $\pm 0.3$ seconds of the labeled P-arrival. False positives ($FP$) are assigned to picks outside the window or to extra picks that are not true positives. False negatives ($FN$) occur if no picks occur within the window. True negatives are not informative in this context and are not recorded. An example of a model prediction featuring classified picks is shown in Figure \ref{pred_example}.
        
        In our display, we consider summary metrics including $Recall = \frac{TP}{TP + FN}$ and $F1 = \frac{2TP}{2TP + FP + FN}$ which are only applied to earthquake waveforms. For noise waveforms, we calculate \textit{noise percent correct} as the number of noise waveforms containing no predicted picks ($TN_{noise}$) divided by the total number of noise waveforms ($n_{noise}$). That is, $Noise\,\%\,Correct = \frac{TN_{noise}}{n_{noise}}$. We recognize that classifying noise waveforms is a relatively trivial task and that analysts may be interested in different metrics beyond the ones analyzed here.   
        
        Evaluating accuracy is typically dependent on the classification results and done using regression-based metrics \cite{zhu_phasenet_2018, munchmeyer_which_2022, kim_development_2023, yu_benchmark_2023}. If a pick is classified as a true positive, then we obtain a residual defined as the difference in time between the labeled P-pick and the predicted P-pick. Because different models can produce different numbers of true positives, typical regression metrics like the root mean square residual (RMSR; the square root of sum of the residuals divided by the number of true positives) fail to convey the full story. Therefore, functional based metrics like cumulative RMSR, defined as RMSR calculated using only the residuals whose magnitudes are below some value, are more informative. 

        We define a different threshold for each model instance by choosing the threshold that maximizes the mean of $F1$ and $Noise\,\%\,Correct$ on the validation set in order to seek a balanced performance between noise and earthquake classes. However, this is an arbitrary decision and the choice is typically dependent on the analyst's goals.

    \subsection{Statistical Framework for Evaluation}
        
        In this section, we present a statistical framework for comparing models in terms of their performance and uncertainty arising from data sampling and model training. The statistical framework can be thought of as a mixed effects model or a completely randomized design with sub-sampling. It allows us to distinguish performance effects with uncertainty between the 3 models and different data training budgets. Furthermore, we can segregate and estimate the two sources of variation with uncertainty.         

        For a given metric that summarizes the outcomes of a model instance, denoted $y_{madi}$, where $y$ is the metric (e.g., $Recall$), we set up the statistical model
        \begin{equation} 
            y_{madi} = \gamma + \mu_m + \alpha_a + \theta_{ma} + \epsilon^{data}_{mad} + \epsilon^{train}_{madi}, 
            \label{eq:1}
        \end{equation}
        where $m = \{1\ldots3\}$ indexes the 3 model training approaches, $a = \{1,\ldots ,5\}$ indexes the 5 quantities of clusters (representing the amount of data) at levels $\{1, 3, 6, 9, 12\}$, $d = \{1,\ldots ,12\}$ indexes the randomly selected cluster set (i.e., the data choice selected to train the model), and $i = \{1,2,3,4\}$ indexes the 4 model initializations. Here, $\gamma$ is the grand mean, $\mu_m$ is the effect of model $m$, $\alpha_a$ is the effect of the $a$'th quantity of clusters, and $\theta_{ma}$ is the interaction term. 
        
\begin{figure} %
            \centering
            \includegraphics[]{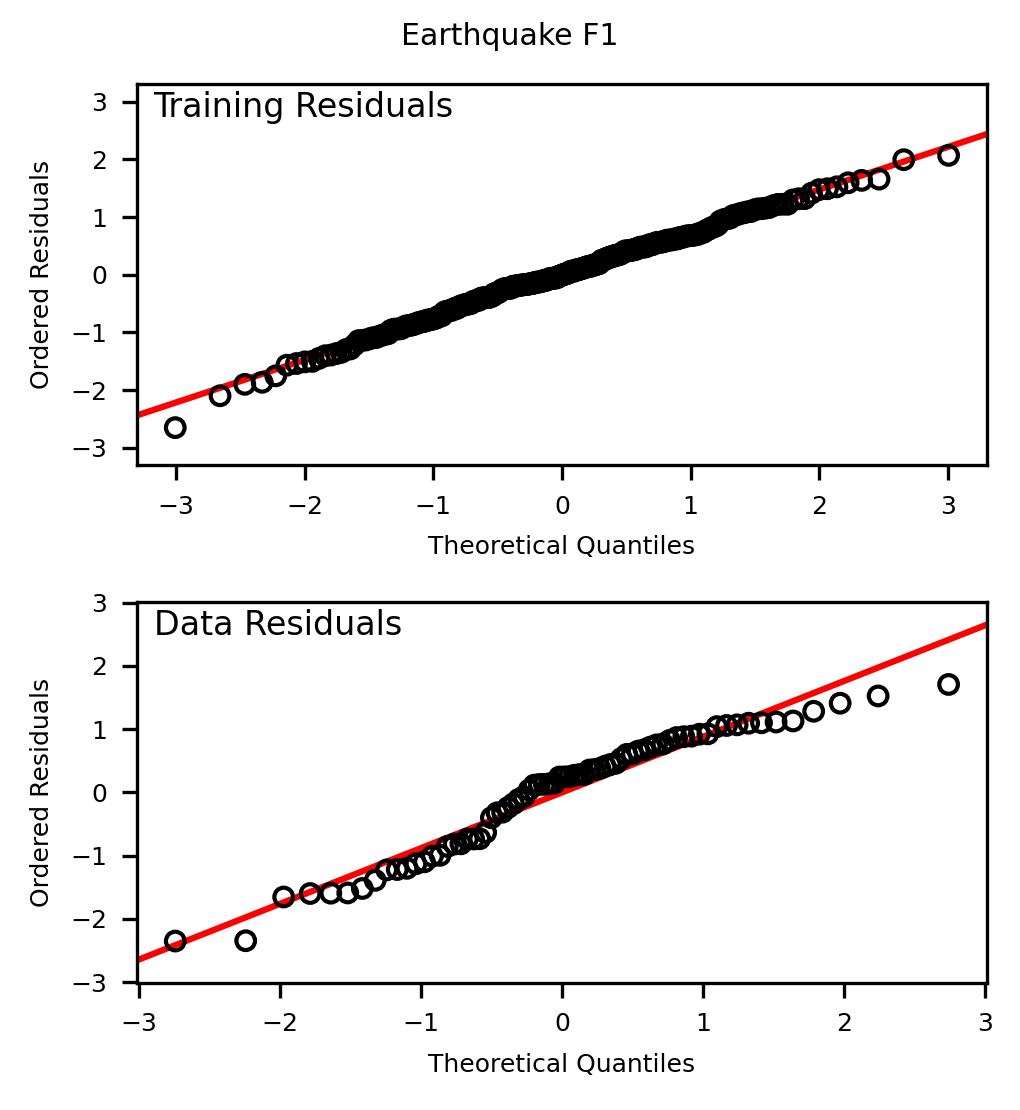}
            \caption{The Gaussian assumptions are visually verified through quantile-quantile plots. As an example, the sorted training (top) and data (bottom) residuals for F1 score are seen to align closely with the theoretical quantiles as defined by the statistical framework as seen in how the points follow the red line.}
            \label{qq}
        \end{figure}

        The uncertainty associated with data sampling and training is segregated into two independent random terms each assuming a zero mean independent and identically distributed Gaussian with variance inherent to a given model and quantity of clusters: $\epsilon^{data}_{mad} \sim \mathcal{N}(0, \sigma^{2 data}_{ma})$ and  $\epsilon^{train}_{madi} \sim \mathcal{N}(0, \sigma^{2 train}_{ma})$. The distribution of the error terms implies a distribution of the metric as $y_{madi} \sim \mathcal{N}(\gamma + \mu_m + \alpha_a + \theta_{ma}, \sigma^{2 data}_{ma} + \sigma^{2 train}_{ma} )$. We recognize that metrics such as recall are bounded between $[0,1]$, so the Gaussian assumption theoretically cannot hold. However, in practice the metrics are calculated from a large number of test waveforms which provides theoretical support for the Gaussian assumption through the central limit theorem. In addition, we find that the variances of these metrics are small compared to the range of the metrics' bounds. Furthermore, these assumptions are reasonable because the Gaussian distribution is capable of approximating confidence even when the error distribution is not strictly Gaussian. In addition, we verified the assumptions through visual inspection of quantile-quantile plots that compare the sorted residuals as a function of the theoretical quantiles for all metrics of interest. An example of this check for F1 score is shown in Figure \ref{qq} demonstrating strong support for the theoretical assumptions. 
        
        Using these models, we estimate the effect of each model and quantity of clusters on the metric by computing the mean over the 12 cluster sets and the 4 initializations. Then, we compute $90\%$ confidence intervals around those estimates using the Gaussian assumptions. Note that the intervals are influenced more heavily by the cluster set (data choice) uncertainty because at each cluster set, we average over the initializations thereby reducing training uncertainty influence.

\section{Results}

    After training the 720 model instances, computing the performance metrics, and organizing them through the statistical framework, we are ready to compare the performance of the models. 

\begin{figure} %
        \centering
        \includegraphics[]{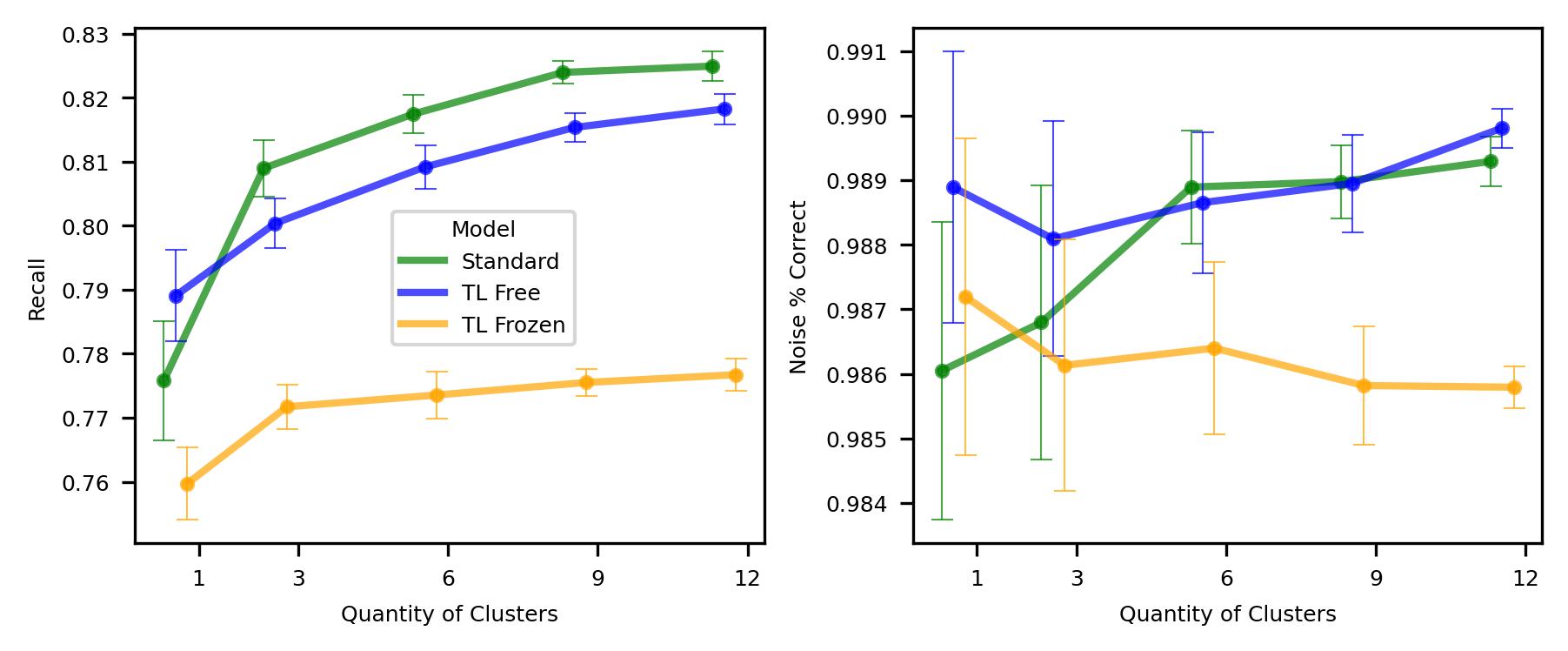}
        \caption{The statistical framework is used to summarize classification results including earthquake recall (left) and noise percent correct (right) as a function of the quantity of clusters used for training for the 3 training approaches. Vertical bars represent 90\% confidence intervals. Largest performance gains occur as the quantity of clusters increases from 1 to 3 and crossing lines indicate different training approaches are superior for different budgets of training data.}
        \label{results}
    \end{figure}

    The classification based summary metrics of recall and noise percent correct are presented in Figure \ref{results} along with 90\% confidence intervals for the Standard, TL Free, and TL Frozen models showing their performance as a function of quantity of clusters used in training. All models improve in recall as more data is added with largest gains between 1 and 3 clusters followed by diminishing returns. The trend is less consistent for noise percent correct where, in contrast to expectations, the TL Frozen model performs slightly worse with increasing data. The uncertainty in performances decreases for all models as more training data is added. 

\begin{figure} %
        \centering
        \includegraphics[]{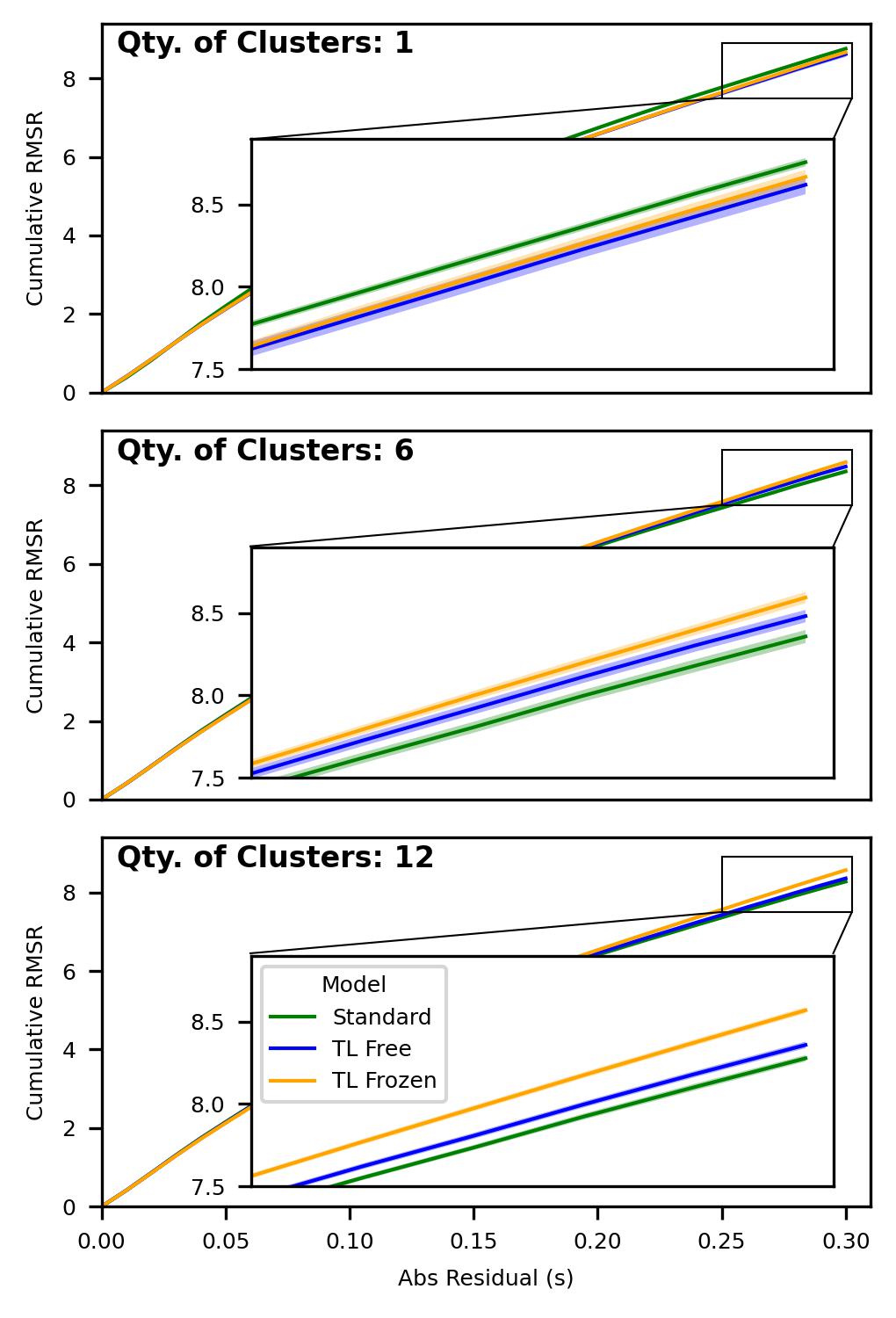}
        \caption{The statistical framework is used to summarize the regression-based functional metric of cumulative RMSR for 1, 6, and 12 clusters (top, middle, and bottom respectively) for the 3 models. Ribbons indicate 90\% confidence intervals. Inset plots zoom into the tails to show how the relative positions of the model curves change as more training data is added indicating that no model is the most efficient learner for all training data budgets.}
        \label{cumulative}
    \end{figure}

    The functional regression based metric of cumulative RMSR demonstrating the evaluation's framework versatility and importance is shown in Figure \ref{cumulative} for the 3 models with 90\% confidence intervals (ribbons) for 1, 6, and 12 quantities of clusters. To compute these, we apply the statistical framework (\ref{eq:1}) treating each discretized element along the domain as a standalone metric. Stark performance differences can be seen between each model especially as more training data is added as seen through tightening confidence intervals. Note that cumulative RMSR ends at $0.3$ seconds because that is the half-width of the true positive window. 

    In both Figures \ref{results} and \ref{cumulative}, we see that there is no clear winner in terms of learning efficiency for all training budgets. The TL Free model demonstrates superiority in the mean performance for recall and cumulative RMSR for the smallest data size of 1 cluster, but is surpassed by the Standard model at 3 clusters and beyond. These results would help practitioners choose the model for their given problem. Depending on the availability of training data and how they prioritize performance metrics, their choice would change. 
    
\begin{figure} %
        \centering
        \includegraphics[]{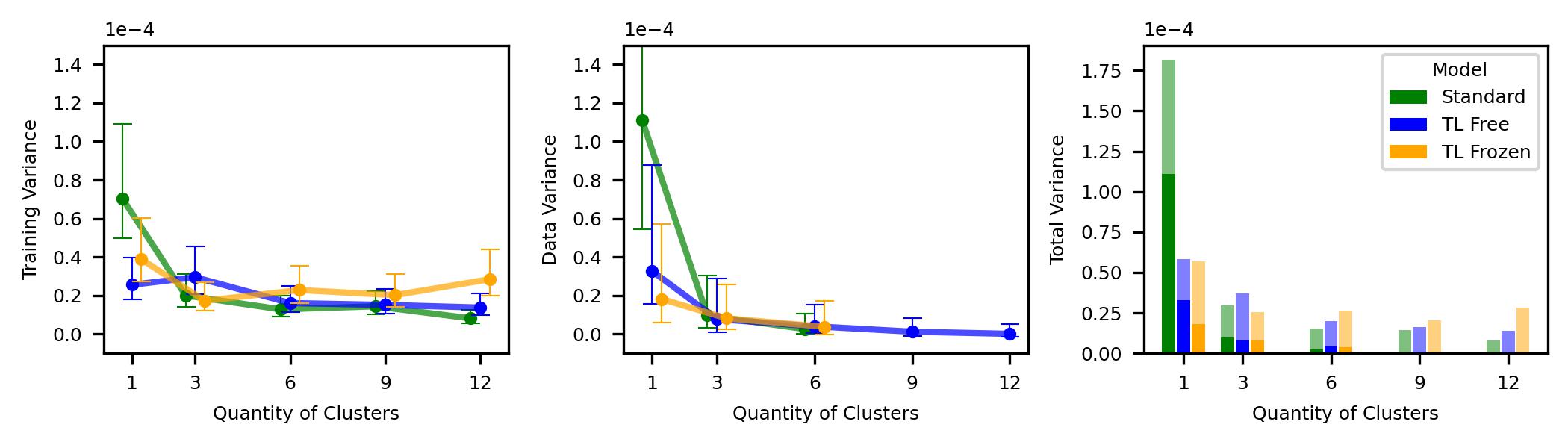}
        \caption{Training variance (left) and data sampling variance (middle) are estimated with 90\% confidence intervals (vertical bars) for F1 for the 3 training approaches. The right panel shows bar plots constructed as the sum of the estimates of training and data variances. The different transparency levels show the contribution of each component where the training contribution is transparent and the data contribution is opaque. Estimating performance uncertainty and attributing the source is informative to practitioners during model selection for setting performance expectations and risk mitigation.}
        \label{variance}
    \end{figure}

    Our evaluation framework allows for estimation of two sources of performance uncertainty including random data sampling and stochastic training processes. As an example, Figure \ref{variance} shows these estimates for F1 for the 3 models across the training data budgets with 90\% confidence intervals. We see that training and data variance decrease as more training data is added and we see some differentiation in the estimated variances for each model. The data variance (middle panel) shrinks quickly towards zero as more data is added and is often estimated as negative. These negative estimates are not shown because negative variance estimates are theoretically impossible. This is a known challenge in these classical statistical frameworks, but negative values essentially mean the data variance component is very small relative to the training variance. 

    The right panel in Figure \ref{variance} shows the relative contributions of each source of uncertainty through bar plots where the opaque and transparent bars are the contributions of data and training variances, respectively. The total variance for all models trends downward with more data with diminishing returns. The uncertainty is mostly dominated by the stochastic training processes when quantities of clusters is 3 or more. Practitioners can use this information to target specific sources of variation that they suspect might dominate their problem, perhaps by running more initializations or strategically labeling data.
    
    We provide an example that emphasizes the risk of evaluating models without considering uncertainty in performance. We consider the scenario where a practitioner is deciding among the 3 training approaches to incorporate into their workflow. They are equipped with a single cluster set, say 1 cluster's worth of data. To make their decision, we imagine that they train one instance of each model using the fixed data. We did exactly this in our training design, but we went further by training each model 4 times under different initializations, not just once. As such, we imagine the practitioner is equally likely to get one of the four recall scores we obtained for each model. Thus, with 3 models and 4 equally likely results for each, there are $4^3=64$ equally likely possible outcomes that the practitioner could realize. If the practitioner ranks each model based on recall score (e.g., first place, second place, third place), due to stochastic training processes, we might expect these rankings to change if the practitioner ran the experiment a second time. The important question then becomes, what is the probability that each model gets each rank?
    
\begin{figure} %
        \centering
        \includegraphics[]{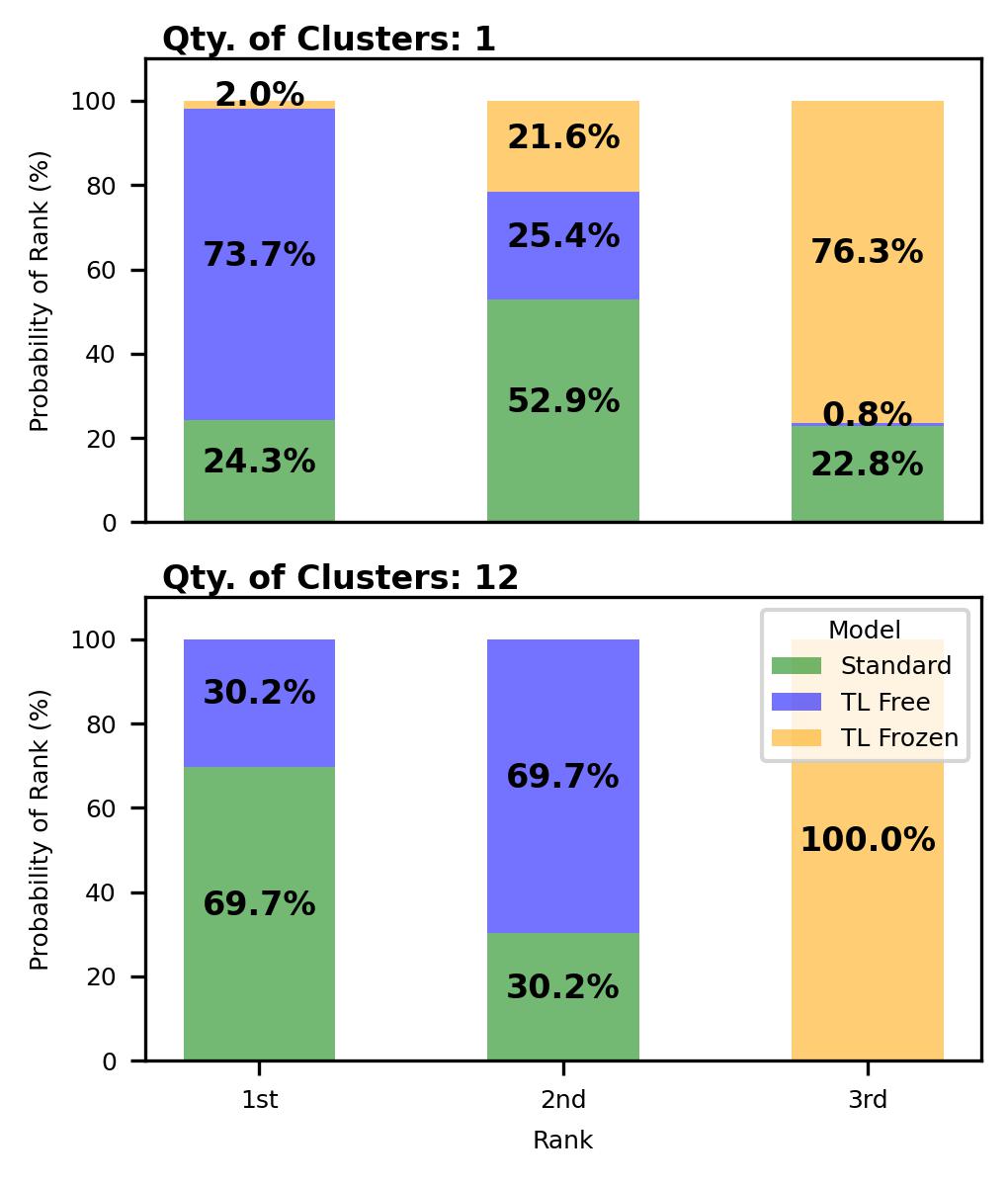}
        \caption{To demonstrate the importance of considering uncertainty in the evaluation framework, the probability of ranking the 3 training approaches based on recall in first, second, or third place is shown for 1 (top) and 12 (bottom) quantities of clusters. Numbers/bars indicate the probability of ranking the associated model in each place if one were to only train one model instance. Numbers less than 100\% indicate that mis-ranking the models is a risk which is detrimental to model selection.}
        \label{rank}
    \end{figure}

    Using our 3 models each initialized 4 times for a fixed quantity of clusters, we calculate the exact probabilities that each model is ranked in first, second, or third place. Furthermore, since we have 12 cluster sets for each quantity of clusters, we conduct these calculation for each of the 12 sets and take the mean to mitigate data bias. The results of this simulation are presented in Figure \ref{rank}, where the probability estimates of ranking each model based on recall in first, second, or third place are shown for 1 and 12 cluster quantities. 
    
    Figure \ref{rank} shows that, due to training and data uncertainty, there is substantial risk of choosing a model that performs more poorly on average. For example, for 1 cluster (top panel), there is a $2.0\%$ chance that TL Frozen has the best recall score. In light of the recall results in Figure \ref{results} (left panel) for 1 cluster, we know that TL Frozen performs substantially worse than the other two models on average. In our case, based purely on recall, it would be a mistake if the practitioner were to choose this model for their workflow. In a similar light, claiming that the TL Frozen model is the state of the art compared to the other two would be a grave mistake as it could encourage technological development in a dead end. Our framework helps guard against such mistakes.

\section{Discussion}

    We present results that will be increasingly relevant for developers of DLMs for the seismic community. In inspecting the results presented in Figure \ref{results}, we see overall poor performance in the TL Frozen model. We hypothesized that freezing half the model at the learned weights from the global dataset would produce the best model for small training data amounts. However, the results negate this hypothesis in our case for datasets comprising 10,000 waveforms (approximately the number of waveforms in a single cluster) or more for the PhaseNet architecture. In a similar vein, we see that models trained from scratch perform better than TL when the data size approaches 30,000 waveforms (3 clusters). This suggests that TL might not always lead to better performance and emphasizes the need to continue comparing TL with traditional training approaches. 

    Our results show that there is no one model that performs the best across all training data quantities. The optimal model choice may depend on the amount of labeled data available for training, especially in low-data scenarios. Another valuable result for the seismic AI community is that expending effort to acquire larger and larger labeled training datasets may be wasteful. This is exhibited by Figure \ref{results} where we see diminishing improvement as the amount of training data increases. This suggest relatively smaller datasets ($\sim$30,000 or $\sim$60,000 waveforms in our case) may be sufficient for reasonable model performance even when training models from scratch. This is good news because large labeled datasets are expensive and sometimes unattainable. We acknowledge that these conclusions may change depending on how a given analyst or developer chooses their data, model architecture, training process, data augmentations, and metric formulations. Estimating learning efficiency could help understand cost and benefit trade off for obtaining more data in this situation.
    
    We emphasized evaluation of models in terms of their learning efficiency while accounting for training and data uncertainty in the metrics of interest. These aspects are valuable for a more wholistic, robust evaluation, but there are more aspects that could supplement our framework by providing context to the raw performance metrics. Some examples of these aspects include the amount of training/computing resources, the intricacy of data augmentations, robustness to noise, predictive uncertainty coverage, architecture complexity, energy efficiency, and model transparency/interpretability. While these important evaluation aspects are not covered in our exposition, our evaluation framework can assist with fair comparison of them going forward.
    
    In this work, we emphasized the importance of controlling data leakage when constructing training splits/budgets to better measure learning efficiency and represent data uncertainty. We constructed our splits by clustering using source locations as a simple and potentially representative approach to how seismic datasets are often geographically constrained. However, clustering on different features informed by domain experts may provide greater control over data leakage, better representation of data sampling procedures, and can target specific performance goals like performance on out-of-distribution test data. 

    The optimal number of clusters is something to be considered alongside estimating data variance. We noted that some data variances are estimated as negative because they are small relative to training variance. One way to more precisely estimate data variance is to increase the number of cluster sets (i.e. from 12 randomly chosen cluster sets to 30). However, we would not see much variation in the cluster sets for large quantities of clusters (e.g., 9 or 12) because we do not sample with replacement. Sampling with replacement would allow larger cluster quantities to have more variety in their cluster compositions. But, this also has a drawback in that some models may get biased towards a specific cluster if it appears multiple times in the cluster set and may not best represent data sampling uncertainty. Another option is to create more clusters such that each cluster set is more likely to be a unique combination of clusters and the probability of a single cluster appearing multiple times is diminished. However, we are limited in decreasing the cluster sizes because as the total number of clusters approaches the number of sources, the issue of data leakage resurfaces. Determining the best solution to this issue is a fruitful topic for future research.

    In our display, we leverage deep ensembles to obtain training uncertainty for performance for a few reasons. Fundamentally, deep ensembles are easy to understand and implement and they only require extending training time. This is in contrast to other uncertainty methods that attempt to estimate prediction uncertainty or approximate the posterior distribution of a given prediction. These methods include Bayesian neural networks \cite{lampinen_bayesian_2001} and approximations thereof including dropout \cite{gal_dropout_2016}, Stochastic Weight Averaging - Gaussian (SWAG) \cite{maddox_simple_2019} and many others \cite{gawlikowski_survey_2022}. However, these approaches enforce distributional assumptions of the model weights, make strong/arbitrary decisions (e.g, dropout), rarely consider different basins of attraction, and are generally more difficult to interpret. Some methods, however, such as MultiSWAG \cite{wilson_bayesian_2022} (see \cite{armstrong_deep-learning_2023} for a seismic based implementation), offer a strong balance between the complex Bayesian assumptions and the simpler empirical ensembles and offer strong promise for estimating performance uncertainty. 
    
    There are also avenues for improvement for the statistical framework. A Bayesian framework could assist when comparing models with similar characteristics through hierarchical sharing of information. In addition, it could also address the issue of estimating negative data variances through constrained priors. 

    Inspired by the remarkable success of foundation models (FMs), a subset of AI (typically DL) models trained on vast amounts of data in a self-supervised fashion that can be adapted for various downstream tasks using relatively smaller labeled datasets \cite{bommasani_opportunities_2022}, in the areas of natural language processing and computer vision, FMs are actively being developed for seismology \cite{sheng_seismic_2025, gao_foundation_2024, liu_seislm_2024,si_seisclip_2024}. As the seismic community prepares to assess the utility of seismic FMs, our work provides some useful directions. Because FMs are pre-trained using a large general dataset and fine-tuned (typically weights for the last couple of layers are set to be updated) using tailored datasets to accomplish downstream tasks, our work highlights the importance of assessing the utility of seismic FMs by comparing them to traditional models. Our results also fortify the importance of learning efficiency as an aspect of future FM evaluation. Since the best model depends on the amount of training data, knowing which model or training approach performs better with less data is valuable information during model selection. And, as our results shows, it may not be a simple one-model-takes-all, but rather a dynamic choice for a given situation.

\section{Conclusion}
    We present an evaluation framework for seismic DLMs that simultaneously considers performance uncertainty and learning efficiency. We demonstrate our evaluation framework using PhaseNet, a popular phase picking DLM to compare 3 training approaches (training from scratch and 2 TL approaches). We implement an extensive training design utilizing deep ensembles and meticulously constructed training splits that partially controls for data leakage. We summarize the resulting 720 models instances through a statistical framework to compare the 3 training approaches across the training budgets with uncertainty.
    
    We demonstrate the effectiveness and versatility of the evaluation framework by comparing classification and functional regression metrics. We show how no model is a global winner in terms of learning efficiency; that it depends on the total amount of training data used. By providing estimates for the data and training uncertainty, a developer or practitioner can better target their efforts to mitigate sources they deem most important for their problem. We show that without the evaluation framework, there is a significant risk of model mis-ranking. Finally, our results demonstrate how TL may not always be the best choice in limited data scenarios, which is particularly important as the seismic community invests in seismic FMs and emphasizes the importance to compare FMs to their simpler DLM counterparts and inspect how training data budgets influence model performance. We hope that this research helps practitioners and developers by providing an evaluation framework for robust model comparison.

\section{Acknowledgements}
    This work was performed under the US Department of Energy, National Nuclear Security Administration’s Office of Defense Nuclear Nonproliferation Research and Development. This work is approved for public release under LA-UR-24-31133. We would like to thank Chengping Chai (Oak Ridge National Laboratory), Lisa Linville (Sandia National Laboratory), and Leanna House (Virginia Tech) for their advice and/or edits.

\section{Appendix}
    \subsection{Data Splits and Cluster Distributions}
    
    Figures A1, A2, A3, \& A4, aim to provide a deeper view into how the clustering impacts the distributions of various signal properties of the waveforms. Since this work focuses primarily on P-picking, we use a 10 second window starting at the labeled P-arrival for these signal properties. In all figures, the testing set is shown in black, and each non-black curve coincides with the identically colored clusters in the main article (Figure 2). All curves are normalized for comparison such that they integrate to 1. A 10 second window was chosen because only about 60\% of the 824,135 waveforms have a labeled S-wave, so an adaptive S-P interval window that captures only P-wave information was infeasible without further analysis. In addition, Figure A5 shows the distribution of S-P intervals for those waveforms that contain labeled S-arrivals. About 87\% of the waveforms have S-P intervals less than 10 seconds, so most or all of the P-wave information is captured in the majority of the waveform, without including S-wave information for particularly long S-P intervals. 

    \renewcommand{\thefigure}{A1}
    \begin{figure} %
        \centering
        \includegraphics[]{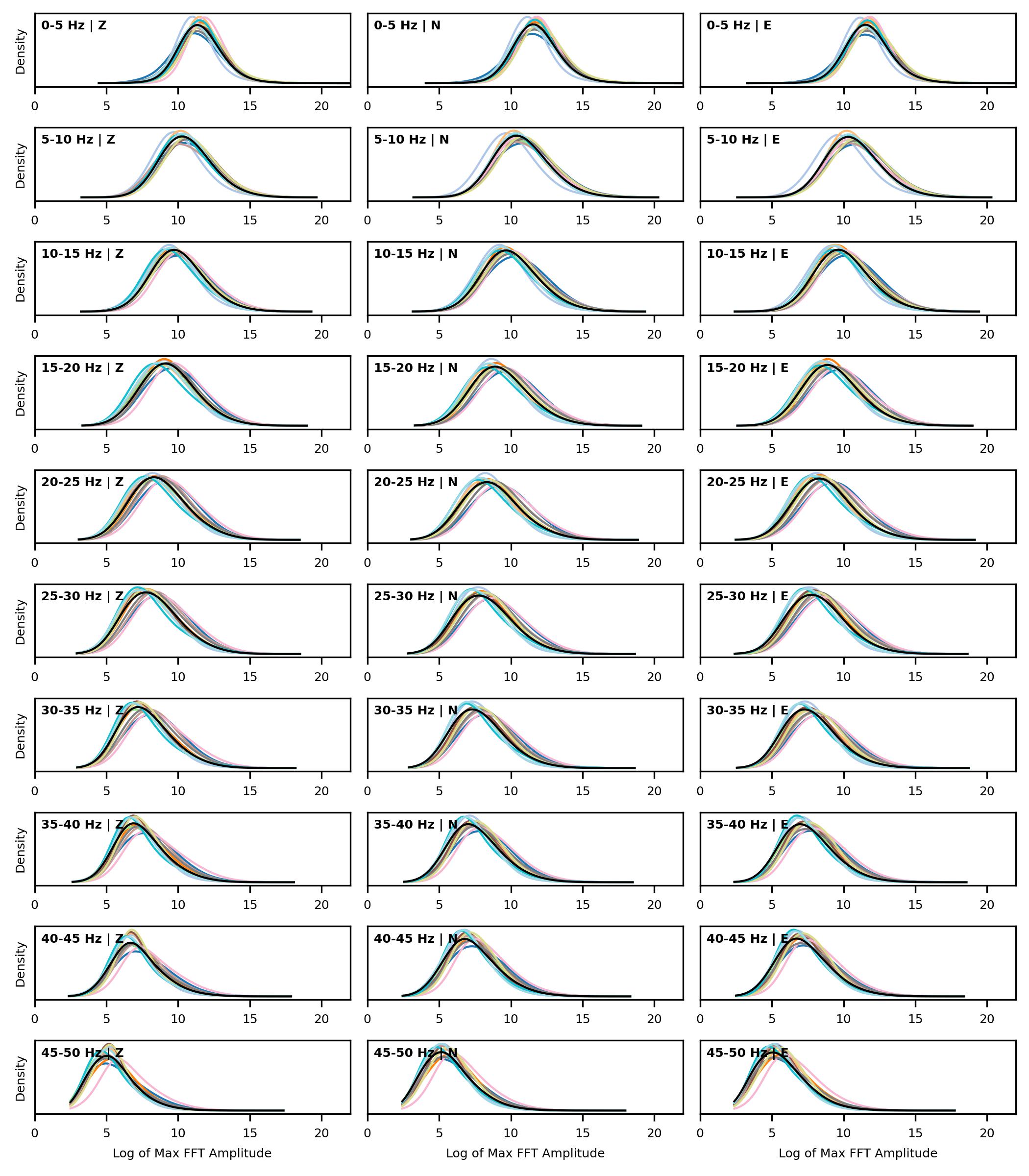}
        \caption{Distributions of the maximum log FFT amplitude within each 5 Hz frequency bin (rows) are shown for the 3 components (columns) for each cluster (non-black curves) and the testing set (black curve).}
        \label{cluster_max_fft}
    \end{figure}

    \renewcommand{\thefigure}{A2}
    \begin{figure} %
        \centering
        \includegraphics[]{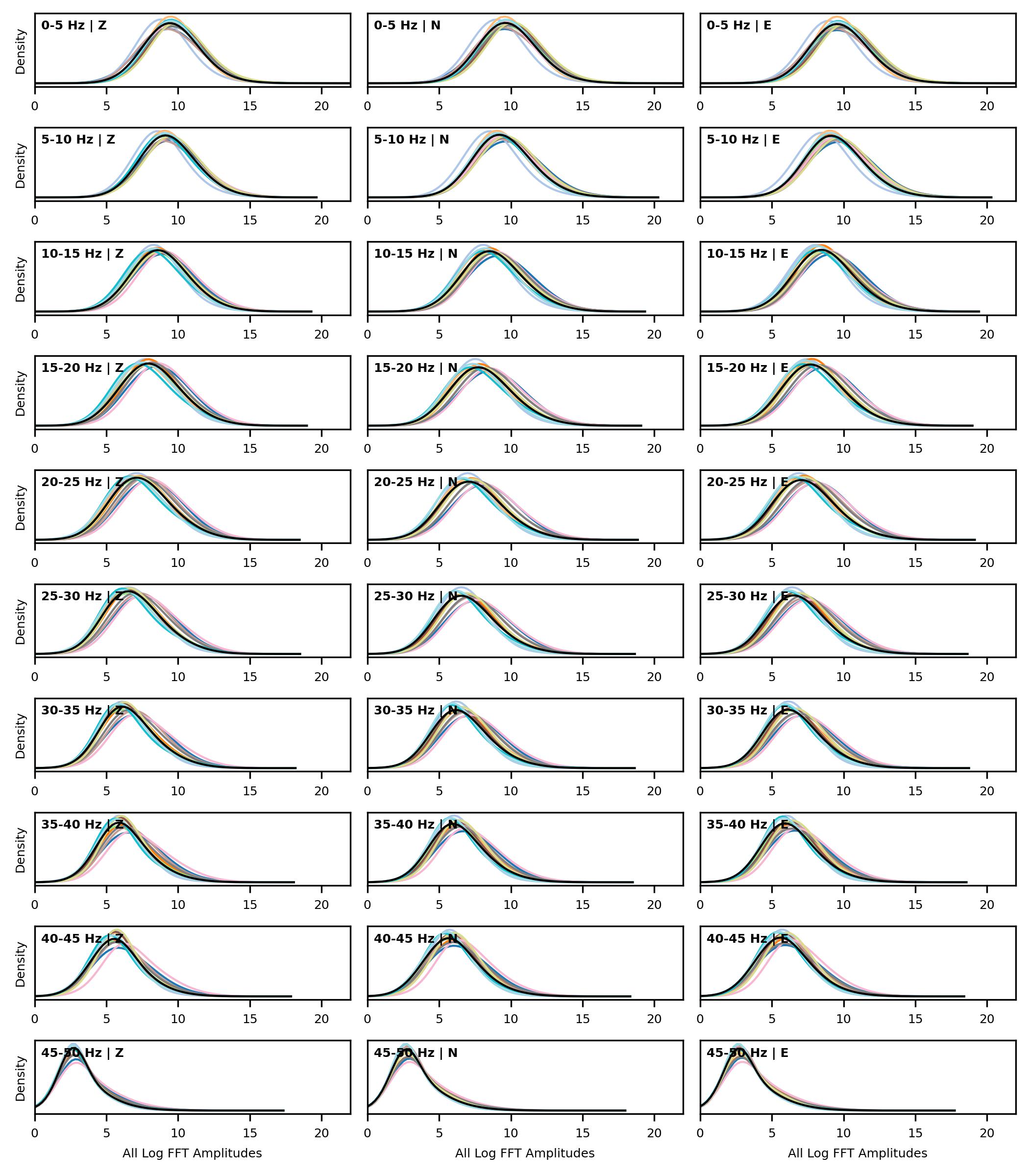}
        \caption{Distributions of all FFT amplitudes within each 5 Hz frequency bin (rows) are shown for the 3 components (columns) for each cluster (non-black curves) and the testing set (black curve).}
        \label{cluster_all_fft}
    \end{figure}

    \renewcommand{\thefigure}{A3}
    \begin{figure} %
        \centering
        \includegraphics[]{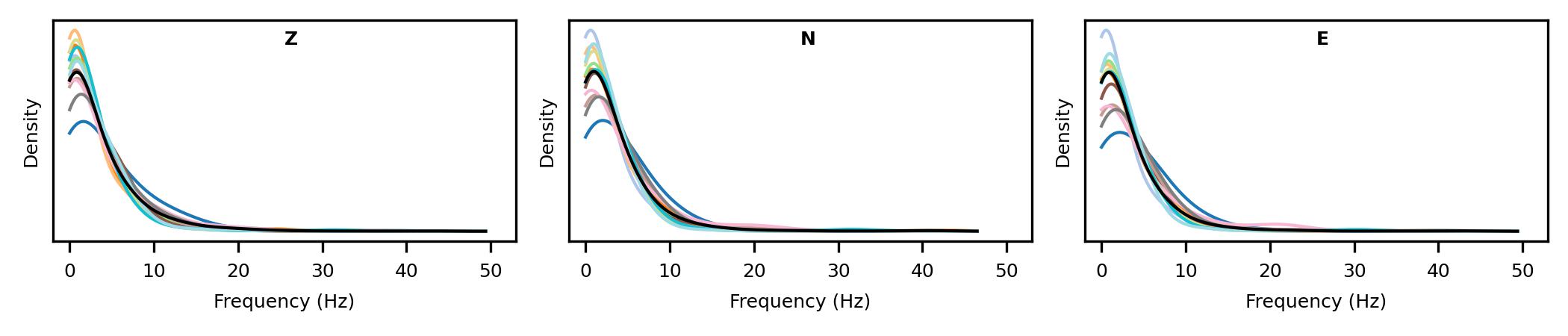}
        \caption{Distributions of the frequencies coinciding with the maximum FFT amplitudes (i.e., the argmax) within a 10 second window starting at the labeled P-arrival are shown for the Z (left), N (middle), and E (right) components for each cluster (non-black curves) and the testing set (black curve).}
        \label{cluster_window_max_freq}
    \end{figure}

    \renewcommand{\thefigure}{A4}
    \begin{figure} %
        \centering
        \includegraphics[]{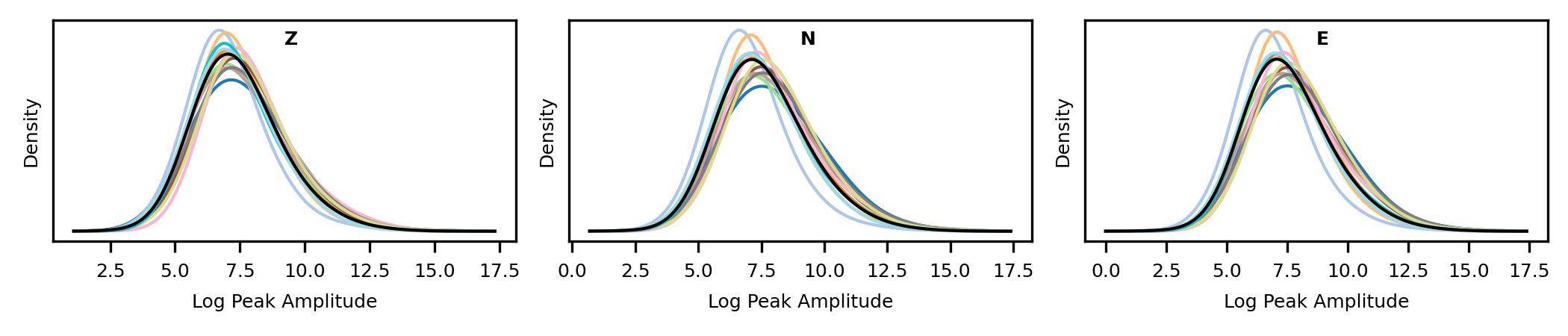}
        \caption{Distributions of the log of the peak amplitudes within a 10 second window starting at the labeled P-arrival are shown for the Z (left), N (middle), and E (right) components for each cluster (non-black curves) and the testing set (black curve).}
        \label{cluster_window_peak_amplitude}
    \end{figure}

    \renewcommand{\thefigure}{A5}
    \begin{figure} %
        \centering
        \includegraphics[]{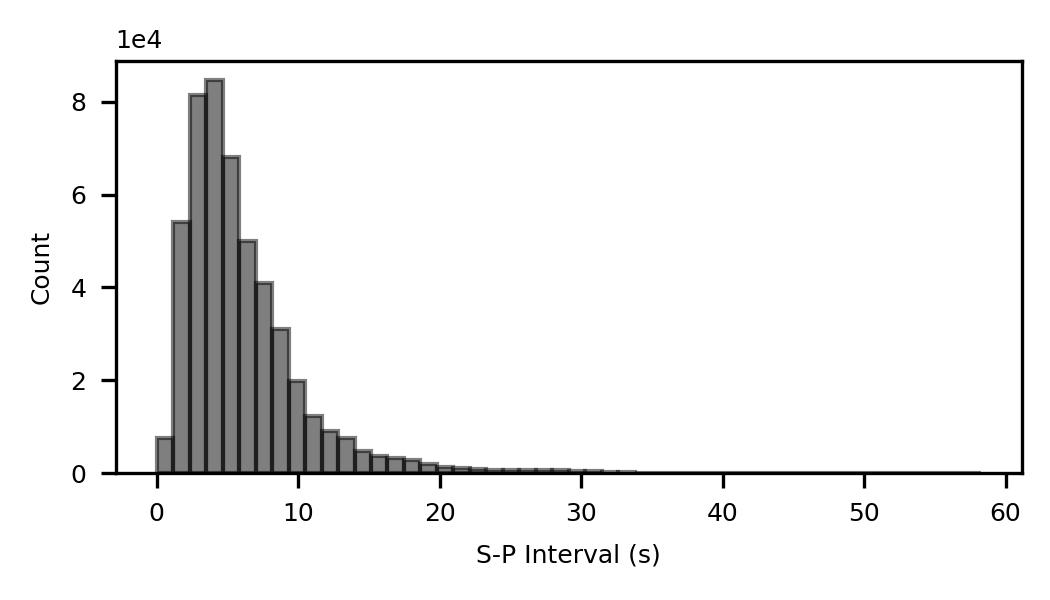}
        \caption{The distribution of the S-P intervals in seconds is shown for the INSTANCE data for which there are labeled S-arrivals. Only about 60\% of the 824,135 waveforms contain a labeled S-arrival.}
        \label{cluster_sp_interval}
    \end{figure}

\newpage

\begin{thebibliography}{10}

\bibitem{zhu_phasenet_2018}
Weiqiang Zhu and Gregory~C Beroza.
\newblock {PhaseNet}: {A} {Deep}-{Neural}-{Network}-{Based} {Seismic} {Arrival}
  {Time} {Picking} {Method}.
\newblock {\em Geophys. J. Int.}, October 2018.

\bibitem{arrowsmith_big_2022}
S.~J. Arrowsmith, D.~T. Trugman, J.~MacCarthy, K.~J. Bergen, D.~Lumley, and
  M.~B. Magnani.
\newblock Big {Data} {Seismology}.
\newblock {\em Rev. Geophys.}, 60(2):e2021RG000769, June 2022.

\bibitem{yu_deep_2021}
Siwei Yu and Jianwei Ma.
\newblock Deep {Learning} for {Geophysics}: {Current} and {Future} {Trends}.
\newblock {\em Rev. Geophys.}, 59(3):e2021RG000742, September 2021.

\bibitem{mousavi_deep-learning_2022}
S.~Mostafa Mousavi and Gregory~C. Beroza.
\newblock Deep-learning seismology.
\newblock {\em Science}, 377(6607):eabm4470, August 2022.

\bibitem{mousavi_applications_2024}
S.~Mostafa Mousavi, Gregory~C. Beroza, Tapan Mukerji, and Majid Rasht-Behesht.
\newblock Applications of deep neural networks in exploration seismology: {A}
  technical survey.
\newblock {\em Geophysics}, 89(1):WA95--WA115, January 2024.

\bibitem{woollam_seisbenchtoolbox_2022}
Jack Woollam, Jannes Münchmeyer, Frederik Tilmann, Andreas Rietbrock, Dietrich
  Lange, Thomas Bornstein, Tobias Diehl, Carlo Giunchi, Florian Haslinger,
  Dario Jozinović, Alberto Michelini, Joachim Saul, and Hugo Soto.
\newblock {SeisBench}—{A} {Toolbox} for {Machine} {Learning} in {Seismology}.
\newblock {\em Seismol. Res. Lett.}, 93(3):1695--1709, May 2022.

\bibitem{munchmeyer_which_2022}
Jannes Münchmeyer, Jack Woollam, Andreas Rietbrock, Frederik Tilmann, Dietrich
  Lange, Thomas Bornstein, Tobias Diehl, Carlo Giunchi, Florian Haslinger,
  Dario Jozinović, Alberto Michelini, Joachim Saul, and Hugo Soto.
\newblock Which {Picker} {Fits} {My} {Data}? {A} {Quantitative} {Evaluation} of
  {Deep} {Learning} {Based} {Seismic} {Pickers}.
\newblock {\em J. Geophys. Res.: Solid Earth}, 127(1):e2021JB023499, January
  2022.

\bibitem{bouthillier_accounting_2021}
Xavier Bouthillier, Pierre Delaunay, Mirko Bronzi, Assya Trofimov, Brennan
  Nichyporuk, Justin Szeto, Naz Sepah, Edward Raff, Kanika Madan, Vikram
  Voleti, Samira~Ebrahimi Kahou, Vincent Michalski, Dmitriy Serdyuk, Tal Arbel,
  Chris Pal, Gaël Varoquaux, and Pascal Vincent.
\newblock Accounting for {Variance} in {Machine} {Learning} {Benchmarks}, March
  2021.
\newblock arXiv:2103.03098.

\bibitem{reimers_reporting_2017}
Nils Reimers and Iryna Gurevych.
\newblock Reporting {Score} {Distributions} {Makes} a {Difference}:
  {Performance} {Study} of {LSTM}-networks for {Sequence} {Tagging}.
\newblock In {\em Proceedings of the 2017 {Conference} on {Empirical} {Methods}
  in {Natural} {Language} {Processing}}, volume Proceedings of the 2017
  Conference on Empirical Methods in Natural Language Processing, pages
  338--348, Copenhagen, Denmark, July 2017. arXiv.

\bibitem{goldstein_league_1996}
Harvey Goldstein and David~J. Spiegelhalter.
\newblock League {Tables} and {Their} {Limitations}: {Statistical} {Issues} in
  {Comparisons} of {Institutional} {Performance}.
\newblock {\em J. R. Stat. Soc. A}, 159(3):385, 1996.

\bibitem{dietterich_approximate_1998}
Thomas~G. Dietterich.
\newblock Approximate {Statistical} {Tests} for {Comparing} {Supervised}
  {Classification} {Learning} {Algorithms}.
\newblock {\em Neural Comput.}, 10(7):1895--1923, October 1998.

\bibitem{dehghani_benchmark_2021}
Mostafa Dehghani, Yi~Tay, Alexey~A. Gritsenko, Zhe Zhao, Neil Houlsby, Fernando
  Diaz, Donald Metzler, and Oriol Vinyals.
\newblock The {Benchmark} {Lottery}, July 2021.
\newblock arXiv:2107.07002.

\bibitem{wein_follow_2023}
Shira Wein, Christopher Homan, Lora Aroyo, and Chris Welty.
\newblock Follow the leader(board) with confidence: {Estimating} p-values from
  a single test set with item and response variance.
\newblock In {\em Findings of the {Association} for {Computational}
  {Linguistics}: {ACL} 2023}, pages 3138--3161, Toronto, Canada, 2023.
  Association for Computational Linguistics.

\bibitem{hlynsson_measuring_2019}
Hlynur~Davíð Hlynsson, Alberto~N. Escalante-B, and Laurenz Wiskott.
\newblock Measuring the {Data} {Efficiency} of {Deep} {Learning} {Methods}.
\newblock In {\em Proceedings of the 8th {International} {Conference} on
  {Pattern} {Recognition} {Applications} and {Methods}}, pages 691--698, 2019.
\newblock arXiv:1907.02549 [cs].

\bibitem{alzubaidi_survey_2023}
Laith Alzubaidi, Jinshuai Bai, Aiman Al-Sabaawi, Jose Santamaría, A.~S.
  Albahri, Bashar Sami~Nayyef Al-dabbagh, Mohammed~A. Fadhel, Mohamed
  Manoufali, Jinglan Zhang, Ali~H. Al-Timemy, Ye~Duan, Amjed Abdullah, Laith
  Farhan, Yi~Lu, Ashish Gupta, Felix Albu, Amin Abbosh, and Yuantong Gu.
\newblock A survey on deep learning tools dealing with data scarcity:
  definitions, challenges, solutions, tips, and applications.
\newblock {\em J. Big Data}, 10(1):46, April 2023.

\bibitem{bornstein_pickblue_2024}
T.~Bornstein, D.~Lange, J.~Münchmeyer, J.~Woollam, A.~Rietbrock, G.~Barcheck,
  I.~Grevemeyer, and F.~Tilmann.
\newblock {PickBlue}: {Seismic} {Phase} {Picking} for {Ocean} {Bottom}
  {Seismometers} {With} {Deep} {Learning}.
\newblock {\em Earth Space Sci.}, 11(1):e2023EA003332, January 2024.

\bibitem{ferrerflorensa_spanseq_2024}
Alfred Ferrer Florensa, Jose Juan Almagro Armenteros, Henrik Nielsen,
  Frank Møller Aarestrup, and Philip Thomas Lanken Conradsen Clausen.
\newblock {SpanSeq}: similarity-based sequence data splitting method for
  improved development and assessment of deep learning projects.
\newblock {\em NAR Genomics Bioinf.}, 6(3):lqae106, July 2024.

\bibitem{ross_generalized_2018}
Zachary~E. Ross, Men‐Andrin Meier, Egill Hauksson, and Thomas~H. Heaton.
\newblock Generalized {Seismic} {Phase} {Detection} with {Deep} {Learning}.
\newblock {\em Bull. Seismol. Soc. Am.}, 108(5A):2894--2901, October 2018.

\bibitem{mousavi_earthquake_2020}
S.~Mostafa Mousavi, William~L. Ellsworth, Weiqiang Zhu, Lindsay~Y. Chuang, and
  Gregory~C. Beroza.
\newblock Earthquake transformer—an attentive deep-learning model for
  simultaneous earthquake detection and phase picking.
\newblock {\em Nat. Commun.}, 11(1):3952, August 2020.

\bibitem{chen_cubenet_2022}
Guoyi Chen and Junlun Li.
\newblock {CubeNet}: {Array}-{Based} {Seismic} {Phase} {Picking} with {Deep}
  {Learning}.
\newblock {\em Seismol. Res. Lett.}, 93(5):2554--2569, September 2022.

\bibitem{zhu_deep_2019}
Lijun Zhu, Zhigang Peng, James McClellan, Chenyu Li, Dongdong Yao, Zefeng Li,
  and Lihua Fang.
\newblock Deep learning for seismic phase detection and picking in the
  aftershock zone of 2008 {M7}.9 {Wenchuan} {Earthquake}.
\newblock {\em Phys. Earth Planet. Inter.}, 293:106261, August 2019.

\bibitem{zhu_endend_2022}
Weiqiang Zhu, Kai~Sheng Tai, S.~Mostafa Mousavi, Peter Bailis, and Gregory~C.
  Beroza.
\newblock An {End}‐{To}‐{End} {Earthquake} {Detection} {Method} for {Joint}
  {Phase} {Picking} and {Association} {Using} {Deep} {Learning}.
\newblock {\em J. Geophys. Res.: Solid Earth}, 127(3):e2021JB023283, March
  2022.

\bibitem{tokuda_seismic-phase_2023}
Tomoki Tokuda and Hiromichi Nagao.
\newblock Seismic-phase detection using multiple deep learning models for
  global and local representations of waveforms.
\newblock {\em Geophys. J. Int.}, 235(2):1163--1182, July 2023.

\bibitem{niksejel_obstransformer_2024}
Alireza Niksejel and Miao Zhang.
\newblock {OBSTransformer}: a deep-learning seismic phase picker for {OBS} data
  using automated labelling and transfer learning.
\newblock {\em Geophys. J. Int.}, 237(1):485--505, February 2024.

\bibitem{cianetti_comparison_2021}
S.~Cianetti, R.~Bruni, S.~Gaviano, D.~Keir, D.~Piccinini, G.~Saccorotti, and
  C.~Giunchi.
\newblock Comparison of {Deep} {Learning} {Techniques} for the {Investigation}
  of a {Seismic} {Sequence}: {An} {Application} to the 2019, {Mw} 4.5 {Mugello}
  ({Italy}) {Earthquake}.
\newblock {\em J. Geophys. Res.: Solid Earth}, 126(12):e2021JB023405, December
  2021.

\bibitem{park_basement_2022}
Yongsoo Park, Gregory~C. Beroza, and William~L. Ellsworth.
\newblock Basement {Fault} {Activation} before {Larger} {Earthquakes} in
  {Oklahoma} and {Kansas}.
\newblock {\em Seism. Rec.}, 2(3):197--206, July 2022.

\bibitem{yoon_detailed_2023}
Clara~E. Yoon, Elizabeth~S. Cochran, Elizabeth~A. Vanacore, Victor Huerfano,
  Gisela Báez-Sánchez, John~D. Wilding, and Jonathan Smith.
\newblock A {Detailed} {View} of the 2020–2023 {Southwestern} {Puerto} {Rico}
  {Seismic} {Sequence} with {Deep} {Learning}.
\newblock {\em Bull. Seismol. Soc. Am.}, 113(6):2377--2415, December 2023.

\bibitem{armstrong_deep-learning_2023}
Alysha~D. Armstrong, Zachary Claerhout, Ben Baker, and Keith~D. Koper.
\newblock A {Deep}-{Learning} {Phase} {Picker} with {Calibrated}
  {Bayesian}-{Derived} {Uncertainties} for {Earthquakes} in the {Yellowstone}
  {Volcanic} {Region}.
\newblock {\em Bull. Seismol. Soc. Am.}, 113(6):2323--2344, December 2023.

\bibitem{mousavi_stanford_2019}
S.~Mostafa Mousavi, Yixiao Sheng, Weiqiang Zhu, and Gregory~C. Beroza.
\newblock {STanford} {EArthquake} {Dataset} ({STEAD}): {A} {Global} {Data}
  {Set} of {Seismic} {Signals} for {AI}.
\newblock {\em IEEE Access}, 7:179464--179476, 2019.

\bibitem{michelini_instance_2021}
Alberto Michelini, Spina Cianetti, Sonja Gaviano, Carlo Giunchi, Dario
  Jozinović, and Valentino Lauciani.
\newblock {INSTANCE} – the {Italian} seismic dataset for machine learning.
\newblock {\em Earth Syst. Sci. Data}, 13(12):5509--5544, November 2021.

\bibitem{pedregosa_scikit-learn_2018}
Fabian Pedregosa, Gaël Varoquaux, Alexandre Gramfort, Vincent Michel, Bertrand
  Thirion, Olivier Grisel, Mathieu Blondel, Andreas Müller, Joel Nothman,
  Gilles Louppe, Peter Prettenhofer, Ron Weiss, Vincent Dubourg, Jake
  Vanderplas, Alexandre Passos, David Cournapeau, Matthieu Brucher, Matthieu
  Perrot, and Édouard Duchesnay.
\newblock Scikit-learn: {Machine} {Learning} in {Python}, June 2018.
\newblock arXiv:1201.0490.

\bibitem{kim_development_2023}
Ahyi Kim, Yuji Nakamura, Yohei Yukutake, Hiroki Uematsu, and Yuki Abe.
\newblock Development of a high-performance seismic phase picker using deep
  learning in the {Hakone} volcanic area.
\newblock {\em Earth Planets Space}, 75(1):85, May 2023.

\bibitem{ronneberger-unet_2015}
Olaf Ronneberger, Philipp Fischer, and Thomas Brox.
\newblock U-{Net}: {Convolutional} {Networks} for {Biomedical} {Image}
  {Segmentation}.
\newblock In {\em Medical {Image} {Computing} and {Computer}-{Assisted}
  {Intervention} – {MICCAI} 2015}, volume 9351, pages 234--241, Cham, 2015.
  Springer International Publishing.
\newblock Series Title: Lecture Notes in Computer Science.

\bibitem{pita-sllim_parametric_2023}
Olivia Pita-Sllim, Calum~J. Chamberlain, John Townend, and Emily Warren-Smith.
\newblock Parametric {Testing} of {EQTransformer}’s {Performance} against a
  {High}-{Quality}, {Manually} {Picked} {Catalog} for {Reliable} and {Accurate}
  {Seismic} {Phase} {Picking}.
\newblock {\em Seism. Rec.}, 3(4):332--341, October 2023.

\bibitem{park_making_2024}
Yongsoo Park, Brent~G. Delbridge, and David~R. Shelly.
\newblock Making {Phase}-{Picking} {Neural} {Networks} {More} {Consistent} and
  {Interpretable}.
\newblock {\em Seism. Rec.}, 4(1):72--80, January 2024.

\bibitem{chai_using_2020}
Chengping Chai, Monica Maceira, Hector~J. Santos‐Villalobos, Singanallur~V.
  Venkatakrishnan, Martin Schoenball, Weiqiang Zhu, Gregory~C. Beroza, Clifford
  Thurber, and {EGS Collab Team}.
\newblock Using a {Deep} {Neural} {Network} and {Transfer} {Learning} to
  {Bridge} {Scales} for {Seismic} {Phase} {Picking}.
\newblock {\em Geophys. Res. Lett.}, 47(16):e2020GL088651, August 2020.

\bibitem{amiri_fine-tuning_2020}
Mina Amiri, Rupert Brooks, and Hassan Rivaz.
\newblock Fine-{Tuning} {U}-{Net} for {Ultrasound} {Image} {Segmentation}:
  {Different} {Layers}, {Different} {Outcomes}.
\newblock {\em IEEE Trans. Ultrason., Ferroelect., Freq. Contr.},
  67(12):2510--2518, December 2020.

\bibitem{lakshminarayanan_simple_2017}
Balaji Lakshminarayanan, Alexander Pritzel, and Charles Blundell.
\newblock Simple and {Scalable} {Predictive} {Uncertainty} {Estimation} using
  {Deep} {Ensembles}, November 2017.
\newblock arXiv:1612.01474.

\bibitem{huang_snapshot_2017}
Gao Huang, Yixuan Li, Geoff Pleiss, Zhuang Liu, John~E. Hopcroft, and Kilian~Q.
  Weinberger.
\newblock Snapshot {Ensembles}: {Train} 1, get {M} for free, March 2017.
\newblock arXiv:1704.00109.

\bibitem{ganaie_ensemble_2022}
M.A. Ganaie, Minghui Hu, A.K. Malik, M.~Tanveer, and P.N. Suganthan.
\newblock Ensemble deep learning: {A} review.
\newblock {\em Eng. Appl. Artif. Intell.}, 115:105151, October 2022.

\bibitem{paszke_pytorch_2019}
Adam Paszke, Sam Gross, Francisco Massa, Adam Lerer, James Bradbury, Gregory
  Chanan, Trevor Killeen, Zeming Lin, Natalia Gimelshein, Luca Antiga, Alban
  Desmaison, Andreas Köpf, Edward Yang, Zach DeVito, Martin Raison, Alykhan
  Tejani, Sasank Chilamkurthy, Benoit Steiner, Lu~Fang, Junjie Bai, and Soumith
  Chintala.
\newblock {PyTorch}: {An} {Imperative} {Style}, {High}-{Performance} {Deep}
  {Learning} {Library}, December 2019.
\newblock arXiv:1912.01703.

\bibitem{kingma_adam_2017}
Diederik~P. Kingma and Jimmy Ba.
\newblock Adam: {A} {Method} for {Stochastic} {Optimization}, January 2017.
\newblock arXiv:1412.6980.

\bibitem{yu_benchmark_2023}
Ziye Yu, Weitao Wang, and Yini Chen.
\newblock Benchmark on the accuracy and efficiency of several neural network
  based phase pickers using datasets from {China} {Seismic} {Network}.
\newblock {\em Earthquake Sci.}, 36(2):113--131, April 2023.

\bibitem{lampinen_bayesian_2001}
Jouko Lampinen and Aki Vehtari.
\newblock Bayesian approach for neural networks—review and case studies.
\newblock {\em Neural Networks}, 14(3):257--274, April 2001.

\bibitem{gal_dropout_2016}
Yarin Gal and Zoubin Ghahramani.
\newblock Dropout as a {Bayesian} {Approximation}: {Representing} {Model}
  {Uncertainty} in {Deep} {Learning}, October 2016.
\newblock arXiv:1506.02142.

\bibitem{maddox_simple_2019}
Wesley Maddox, Timur Garipov, Pavel Izmailov, Dmitry Vetrov, and Andrew~Gordon
  Wilson.
\newblock A {Simple} {Baseline} for {Bayesian} {Uncertainty} in {Deep}
  {Learning}, December 2019.
\newblock arXiv:1902.02476.

\bibitem{gawlikowski_survey_2022}
Jakob Gawlikowski, Cedrique Rovile~Njieutcheu Tassi, Mohsin Ali, Jongseok Lee,
  Matthias Humt, Jianxiang Feng, Anna Kruspe, Rudolph Triebel, Peter Jung,
  Ribana Roscher, Muhammad Shahzad, Wen Yang, Richard Bamler, and Xiao~Xiang
  Zhu.
\newblock A {Survey} of {Uncertainty} in {Deep} {Neural} {Networks}, January
  2022.
\newblock arXiv:2107.03342.

\bibitem{wilson_bayesian_2022}
Andrew~Gordon Wilson and Pavel Izmailov.
\newblock Bayesian {Deep} {Learning} and a {Probabilistic} {Perspective} of
  {Generalization}, March 2022.
\newblock arXiv:2002.08791.

\bibitem{bommasani_opportunities_2022}
Rishi Bommasani, Drew~A. Hudson, Ehsan Adeli, Russ Altman, Simran Arora, Sydney
  von Arx, Michael~S. Bernstein, Jeannette Bohg, Antoine Bosselut, Emma
  Brunskill, Erik Brynjolfsson, Shyamal Buch, Dallas Card, Rodrigo Castellon,
  Niladri Chatterji, Annie Chen, Kathleen Creel, Jared~Quincy Davis, Dora
  Demszky, Chris Donahue, Moussa Doumbouya, Esin Durmus, Stefano Ermon, John
  Etchemendy, Kawin Ethayarajh, Li~Fei-Fei, Chelsea Finn, Trevor Gale, Lauren
  Gillespie, Karan Goel, Noah Goodman, Shelby Grossman, Neel Guha, Tatsunori
  Hashimoto, Peter Henderson, John Hewitt, Daniel~E. Ho, Jenny Hong, Kyle Hsu,
  Jing Huang, Thomas Icard, Saahil Jain, Dan Jurafsky, Pratyusha Kalluri,
  Siddharth Karamcheti, Geoff Keeling, Fereshte Khani, Omar Khattab, Pang~Wei
  Koh, Mark Krass, Ranjay Krishna, Rohith Kuditipudi, Ananya Kumar, Faisal
  Ladhak, Mina Lee, Tony Lee, Jure Leskovec, Isabelle Levent, Xiang~Lisa Li,
  Xuechen Li, Tengyu Ma, Ali Malik, Christopher~D. Manning, Suvir Mirchandani,
  Eric Mitchell, Zanele Munyikwa, Suraj Nair, Avanika Narayan, Deepak
  Narayanan, Ben Newman, Allen Nie, Juan~Carlos Niebles, Hamed Nilforoshan,
  Julian Nyarko, Giray Ogut, Laurel Orr, Isabel Papadimitriou, Joon~Sung Park,
  Chris Piech, Eva Portelance, Christopher Potts, Aditi Raghunathan, Rob Reich,
  Hongyu Ren, Frieda Rong, Yusuf Roohani, Camilo Ruiz, Jack Ryan, Christopher
  Ré, Dorsa Sadigh, Shiori Sagawa, Keshav Santhanam, Andy Shih, Krishnan
  Srinivasan, Alex Tamkin, Rohan Taori, Armin~W. Thomas, Florian Tramèr,
  Rose~E. Wang, William Wang, Bohan Wu, Jiajun Wu, Yuhuai Wu, Sang~Michael Xie,
  Michihiro Yasunaga, Jiaxuan You, Matei Zaharia, Michael Zhang, Tianyi Zhang,
  Xikun Zhang, Yuhui Zhang, Lucia Zheng, Kaitlyn Zhou, and Percy Liang.
\newblock On the {Opportunities} and {Risks} of {Foundation} {Models}, July
  2022.
\newblock arXiv:2108.07258.

\bibitem{sheng_seismic_2025}
Hanlin Sheng, Xinming Wu, Xu~Si, Jintao Li, Sibo Zhang, and Xudong Duan.
\newblock Seismic foundation model: {A} next generation deep-learning model in
  geophysics.
\newblock {\em Geophysics}, 90(2):IM59--IM79, March 2025.

\bibitem{gao_foundation_2024}
Hang Gao, Xinming Wu, Luming Liang, Hanlin Sheng, Xu~Si, Gao Hui, and Yaxing
  Li.
\newblock A foundation model enpowered by a multi-modal prompt engine for
  universal seismic geobody interpretation across surveys, September 2024.
\newblock arXiv:2409.04962.

\bibitem{liu_seislm_2024}
Tianlin Liu, Jannes Münchmeyer, Laura Laurenti, Chris Marone, Maarten V.~de
  Hoop, and Ivan Dokmanić.
\newblock {SeisLM}: a {Foundation} {Model} for {Seismic} {Waveforms}, October
  2024.
\newblock arXiv:2410.15765.

\bibitem{si_seisclip_2024}
Xu~Si, Xinming Wu, Hanlin Sheng, Jun Zhu, and Zefeng Li.
\newblock {SeisCLIP}: {A} {Seismology} {Foundation} {Model} {Pre}-{Trained} by
  {Multimodal} {Data} for {Multipurpose} {Seismic} {Feature} {Extraction}.
\newblock {\em IEEE Trans. Geosci. Remote Sens.}, 62:1--13, 2024.

\end{thebibliography}

\end{document}